\documentclass{article} 
\usepackage{iclr2027_conference,times}

\usepackage{amsmath,amsfonts,bm}

\def\eqref#1{equation~\ref{#1}}

\def\1{\bm{1}}

\DeclareMathAlphabet{\mathsfit}{\encodingdefault}{\sfdefault}{m}{sl}
\SetMathAlphabet{\mathsfit}{bold}{\encodingdefault}{\sfdefault}{bx}{n}

\usepackage{hyperref}
\usepackage{url}
\usepackage{graphicx}
\usepackage{hyperref}
\usepackage{caption}
\usepackage{booktabs, array, wrapfig}
\usepackage{multirow, multicol, makecell, colortbl}
\usepackage{bm}

\title{Unified Generative Video Compression with Autoregressive and Bidirectional\\Video Diffusion}
\author{%
\parbox[t]{\dimexpr\textwidth-2\tabcolsep\relax}{\raggedright
\mbox{\textbf{Tianyu Zhang}$^{1}$}\quad
\mbox{\textbf{Shuai Huo}$^{2}$}\quad
\mbox{\textbf{Dengchao Jin}$^{2}$}\quad
\mbox{\textbf{Hewei Liu}$^{2}$}\quad
\mbox{\textbf{Meng Lei}$^{2}$}\quad
\mbox{\textbf{Tianwen Wu}$^{2}$}\quad
\mbox{\textbf{Jiawen Gu}$^{2}$}\quad
\mbox{\textbf{Chao Zhou}$^{2}$}\quad
\mbox{\textbf{Li Li}$^{1}$}\quad
\mbox{\textbf{Dong Liu}$^{1}$} \\[5pt]
\textnormal{$^{1}$University of Science and Technology of China}\quad \textnormal{$^{2}$Kuaishou Technology}
}%
}

\iclrfinalcopy 

\begin{document}

\maketitle
\lhead{Preprint}

\begin{abstract}
Video diffusion models can reconstruct visually compelling content from severely compressed representations at ultra-low bitrates, but they impose a rigid deployment trade-off. Bidirectional (BI) models achieve high quality through dense full-sequence attention, yet require the entire sequence before reconstruction and incur substantial latency. Autoregressive (AR) models enable low-delay streaming, but their restricted temporal context reduces compression efficiency. This tension is particularly relevant when a video stream must be decoded in real time yet is also stored for later, quality-oriented playback, making AR and BI models natural choices for the live and replay stages, respectively. To support both stages, we present \textbf{UniGVC}, a \textbf{Uni}fied \textbf{G}enerative \textbf{V}ideo \textbf{C}odec that supports two decoding modes from a single bitstream. UniGVC-BI employs a BI-DiT with dense attention for quality-oriented random-access decoding, whereas UniGVC-AR uses windowed AR-DiT for low-delay streaming. Both paths share a hierarchical latent codec that recurrently produces the same conditions with ultra-low bitrates, allowing UniGVC-BI to serve as a conditional teacher for single-step distillation of UniGVC-AR. A tiny conditional video decoder further reduces streaming latency while preserving most of the reconstruction quality. On standard 1080p benchmarks, UniGVC-BI achieves the best overall rate-distortion-perception trade-off among the evaluated methods and produces temporally stable video at rates as low as 0.0005 bits per pixel. Meanwhile, UniGVC-AR remains competitive while decoding 1080p video at 36.85 FPS, over $250\times$ faster than UniGVC-BI. Code and models will be released later.
\end{abstract}

\section{Introduction}
\label{sec:introduction}

Video compression expects to preserve convincing content at rates where accurate transmission of pixels is impossible. Traditional standards such as H.266/VVC~\citep{bross2021overview}, as well as distortion-optimized neural  codecs~\citep{DCVC,DCVC-FM,DCVC-RT}, are highly effective at removing spatial and temporal redundancy. At low bitrates, however, minimizing pixel distortion encourages averaging over details, producing blur and erasing fine structure. Generative codecs~\citep{qi2025generative,glvc2025,gnvcvd2025} instead accept synthesized but visually consistent details and target the rate-distortion-perception trade-off~\citep{blau2019rethinking}. 

Video diffusion transformer (DiT) can jointly reason about appearance and motion and is consequently well suited to coherent perceptual reconstruction~\citep{DiT,wan2025wan}. Nevertheless, its advantages come at a substantial systems cost: dense bidirectional attention requires the whole coded segment, iterative denoising invokes the large model many times, and the accompanying 3D VAE decoder can itself dominate latency. Such a diffusion decoder is attractive for quality-oriented, delay-tolerant operation, but unsuitable for streaming scenarios.

Autoregressive video diffusion removes future evidence, while causal attention permits a frame or latent chunk to be emitted as soon as its condition arrives, and cached keys and values avoid recomputing the past~\citep{yin2025causvid,huang2025selfforcing}. This distinction is important in applications such as video conference and online teaching, where the same stream must be viewed in real time and retained for later playback. A causal AR decoder can serve the live session with low delay. Once stored, a BI decoder can exploit both past and future context for higher-quality replay. However, existing generative video codecs typically bind the bitstream to one synthesis path.

We explore whether one compressed representation can instead support both regimes. Unlike video generation, the diffusion now receives a strong, temporally aligned codec signal reconstructed from the bitstream. This signal can serve as the autoregressive context. Conditioning on codec features rather than previously reconstructed pixels or clean latents permits parallel training with a shared diffusion timestep while retaining causal, cached inference. The same condition remains compatible with a bidirectional generator, allowing both diffusions to optimize jointly with one shared codec.

\begin{figure*}[t]
    \centering
    \includegraphics[width=\textwidth]{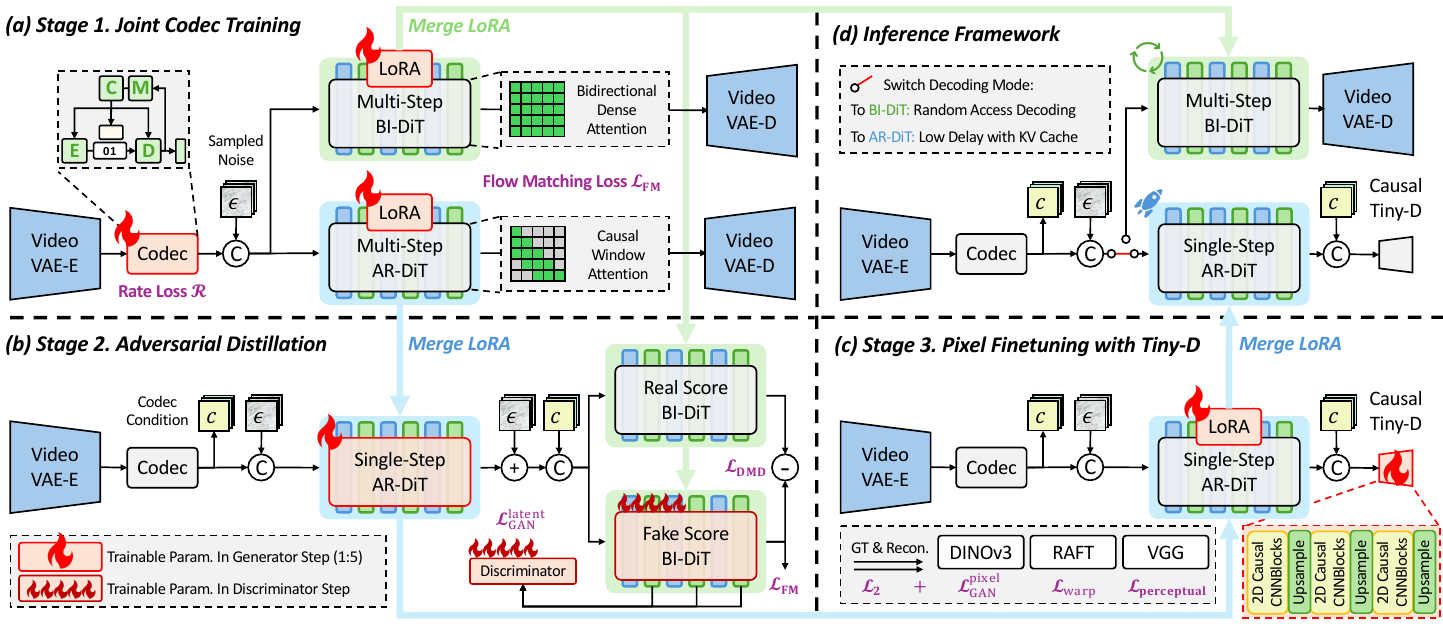}
    \caption{\textbf{Training and inference pipeline of UniGVC.} \textbf{(a)} Training stage 1 jointly adapts a shared latent codec and LoRA modules for a dense bidirectional DiT (BI-DiT) and a windowed causal DiT (AR-DiT). \textbf{(b)} Stage 2 merges the LoRAs and distills AR-DiT to one step using BI-DiT as the real score, fake score, and adversarial supervisor under the same codec condition. \textbf{(c)} Stage 3 refines AR-DiT and a conditional tiny decoder in pixel space. \textbf{(d)} At inference, the same codec condition selects either BI-DiT for random-access decoding or one-step AR-DiT for low-delay streaming.}
    \label{fig:pipeline}
\end{figure*}

Based on this observation, we introduce \textbf{UniGVC} (Fig.~\ref{fig:pipeline}), a unified generative video compression framework at ultra-low bitrates supporting bidirectional and autoregressive diffusion decoding in a continuous video latent space~\citep{wan2025wan}. We facilitate a recurrent hierarchical codec to compress video latents with ultra-low bitrates. The bitstream reconstructs a sequence of shared codec conditions. UniGVC-BI processes those conditions with dense bidirectional attention and iterative denoising, maximizing perceptual quality and temporal consistency for buffered random-access decoding. UniGVC-AR instead uses contextual autoregression and a windowed KV cache for low-delay causal streaming. We distill it into one step with the bidirectional teacher through distribution matching and adversarial supervision, then replace the costly 3D video VAE decoder with a condition-aware tiny causal decoder. Inference consequently supports switching between generative decoders for different decoding quality and latency with the same bits.

We summarize our contributions as:
\begin{itemize}
    \item We present a unified generative video compression framework whose single bitstream supports both a quality-oriented bidirectional decoding mode and a real-time autoregressive decoding mode, covering random-access and low-delay use cases at ultra-low bitrates.
    \item We develop a hierarchical latent video codec for efficient bits allocation at severe compression. To enable autoregressive streaming, we distill AR-DiT into one step exploiting BI-DiT as the distribution teacher under the same codec condition. A conditional tiny decoder further breaks the real-time decoding bottleneck while maintaining most of the quality.
    \item Extensive experiments on video benchmarks demonstrate that UniGVC-BI achieves superior rate-distortion-perception trade-off and temporal stability at bitrates as low as 0.0005 bpp, while UniGVC-AR remains competitive and reaches 36.85 decoding FPS at 1080p.
\end{itemize}

\section{Unified Generative Video Compression}
\label{sec:method}

\begin{figure*}[t]
    \centering
    \includegraphics[width=\textwidth]{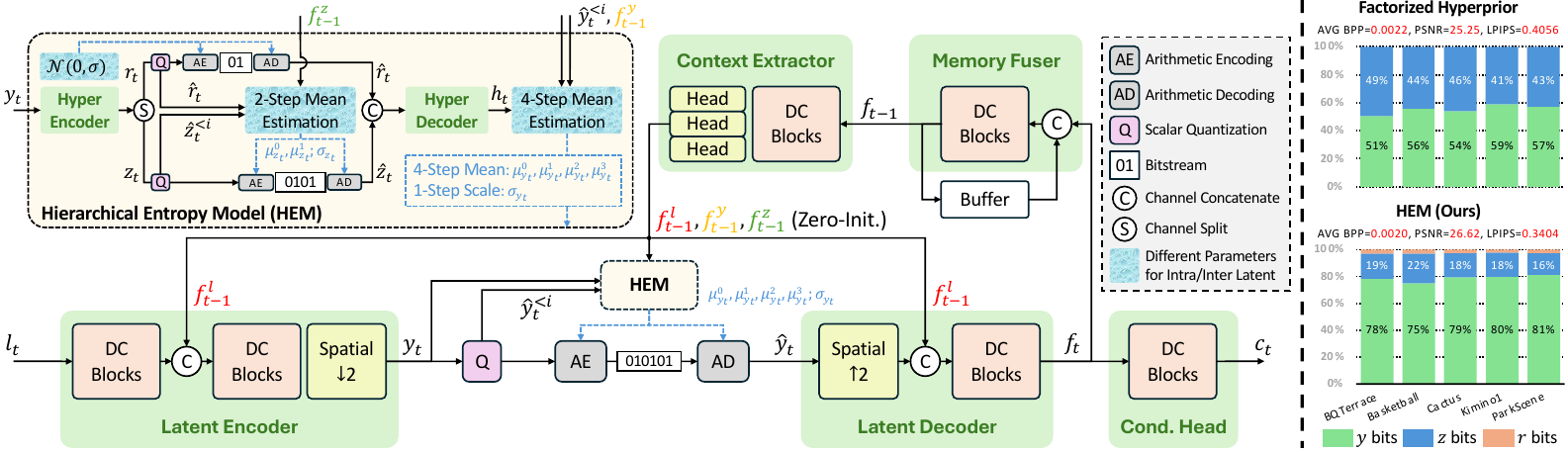}
    \caption{\textbf{Codec overview and analysis on bits allocation.} \textbf{Left:} The unified latent video codec compresses video latent $l_t$ with hierarchical entropy modeling, updates recurrent feature memories, then synthesizes the shared condition $c_t$ for later autoregressive or bidirectional video generation. Most modules are shared across intra and inter latent frames. ``DC Blocks" stands for depthwise convolution blocks~\citep{dcvcuf2026}. \textbf{Right:} Bits allocation comparison on HEVC-B. ``Factorized Hyperprior" omits the $z_t$-$r_t$ decomposition and applies the standard factorized entropy bottleneck, producing $z_t$ with substantial bits without context modeling. In contrast, HEM assigns most bits to the context-modeled $y_t$ and $z_t$, leaving less than $3\%$ bits on weakly modeled $r_t$ with improved compression efficiency and reconstruction quality.}
    \label{fig:codec}
\end{figure*}

\subsection{Hierarchical Latent Video Codec for Ultra-Low Bitrates}
\label{sec:codec}

UniGVC facilitates the temporal compression of the video VAE~\citep{wan2025wan}. Given a video $x\in\mathbb R^{T\times H\times W}$, the VAE encoder maps it to a continuous latent sequence $l$ with 16 channels, $8\times$ spatial downsampling, and $4\times$ temporal downsampling (with a separately represented first frame), making latent coding substantially cheaper than pixel-domain coding. As overviewed in Fig.~\ref{fig:codec}, we construct a recurrent codec unifying intra and inter latent coding to meet the autoregressive streaming. A hierarchical entropy model is introduced to refine bits allocation under ultra-low bitrates.

\textbf{Unified Recurrent Latent Coding.} For latent frame $l_t$, the latent encoder produces a primary latent $y_t$ using the previous decoded feature $f_{t-1}^{l}$. The latent decoder reconstructs a feature $f_t$, updates a recurrent memory buffer~\citep{glvc2025}, and projects it to the condition $c_t$. For the first latent frame, temporal buffer and inputs are zero initialized and a small set of intra-specific entropy layers is activated. This unified intra/inter design keeps every $c_t$ in the same feature space and makes joint optimization with both diffusion decoders possible.

\textbf{Hierarchical Entropy Model.} At extreme rates, conventional factorized hyperprior~\citep{balle2018variational} typically consumes a considerable part of the bit budget (Fig.~\ref{fig:codec} Right). We therefore decompose a residual $r_t$ from the hyper latent $z_t$ that is quantized then entropy-coded with a zero-mean Gaussian and learned channel-wise scales. Conditioned on $\hat r_t$ and temporal feature $f^z_{t-1}$, $\hat z_t$ is coded using a two-pass checkerboard partition~\citep{he2021checkerboard}. The hyper decoder reconstructs the hyperprior from $\left[\hat z_t;\hat r_t\right]$. With temporal feature $f^y_{t-1}$, $\hat y_t$ is modeled by a four-pass quadtree partition~\citep{DCVC-DC}. The resulting hierarchical entropy factorization and the expected bitrate are
\begin{align}
 p(\hat y_t,\hat z_t,\hat r_t\mid f_{t-1}^y,f_{t-1}^z)
 &=p(\hat r_t)
 \prod_{i=1}^{2}p\!\left(\hat z_t^{i}\mid \hat z_t^{<i},\hat r_t,f^z_{t-1}\right)
 \prod_{i=1}^{4}p\!\left(\hat y_t^{i}\mid \hat y_t^{<i},\hat h_t,f^y_{t-1}\right),
 \label{eq:hem-factorization} \\
 \mathcal{R}=\frac{1}{THW}\sum_t& \mathbb{E}\!\left[-\log_2 p(\hat r_t)-\log_2 p(\hat z_t\mid\cdot)-\log_2 p(\hat y_t\mid\cdot)\right].
 \label{eq:rate}
\end{align}
The $r$ branch can now shrink toward zeros, allowing more bits to $y$ and $z$ that are modeled by stronger spatial and temporal context under different feature resolutions. Following DCVC-UF~\citep{dcvcuf2026}, we predict the scales of all partitions in single evaluation and refine only their means across multiple checkerboard or quadtree passes. Arithmetic decoding therefore interacts with accelerators once at each of the $r_t$, $z_t$, and $y_t$ levels, while progressive mean estimation remains on the device.

\textbf{Codec Initialization.} Before incorporating diffusion, we pretrain the codec with
\begin{equation}
 \mathcal{L}_{\mathrm{codec}}
 =\lambda\cdot\mathcal{R} + \sum_t \omega_t\cdot\|l_t-c_t\|_2^2,
 \label{eq:codec-pretrain}
\end{equation}
where $\omega_t$ implements hierarchical frame-quality weighting~\citep{DCVC-FM,glvc2025}. This stage supplies a stable, rate-efficient condition before the large generative models are attached.

\subsection{Joint Training with Contextual Autoregressive Video Diffusion}
\label{sec:contextual-ar}

\begin{figure*}[t]
    \centering
    \includegraphics[width=\textwidth]{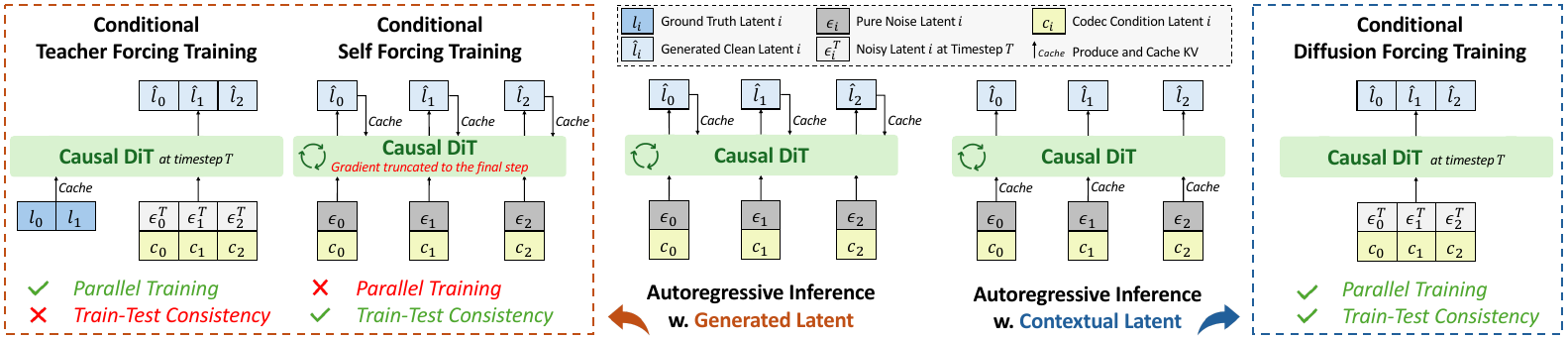}
    \caption{\textbf{Codec-conditioned autoregressive diffusion training and inference.} Autoregression through generated latents requires either teacher forcing (TF), which creates exposure bias, or self forcing (SF), which serializes training. UniGVC instead autoregresses through decoded codec conditions. Conditional diffusion forcing (DF) uses a shared diffusion timestep across frames, permits parallel training, and presents the same bitstream-derived condition at training and inference.}
    \label{fig:ar}
\end{figure*}

\begin{figure*}[t]
\centering
\begin{minipage}[t]{0.491\textwidth}
    \centering
    \includegraphics[width=\linewidth]{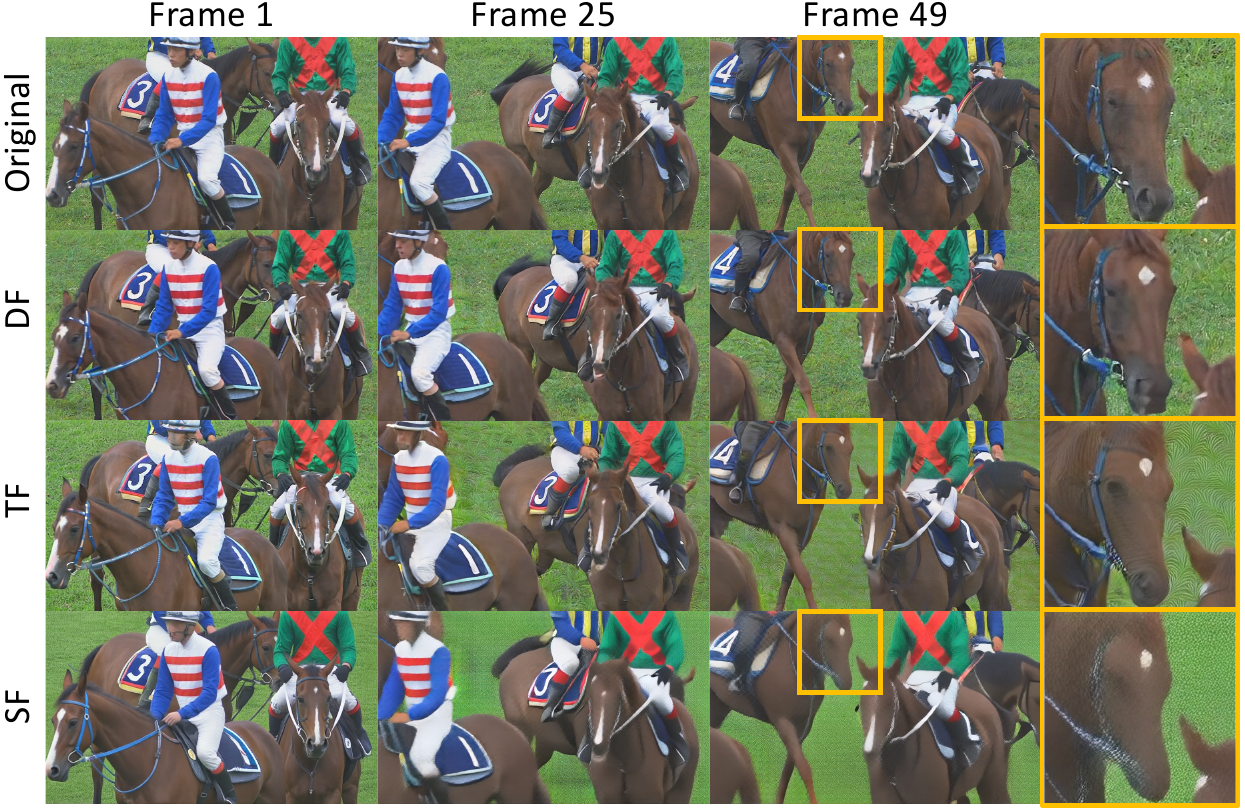}
    \caption{\textbf{Visual analysis of different codec-conditioned AR-DiT.} Videos are tested in 480p and 0.0045-0.005 bpp.}
    \label{fig:ar_visual}
\end{minipage}\hfill
\begin{minipage}[t]{0.49\textwidth}
    \centering
    \includegraphics[width=\linewidth]{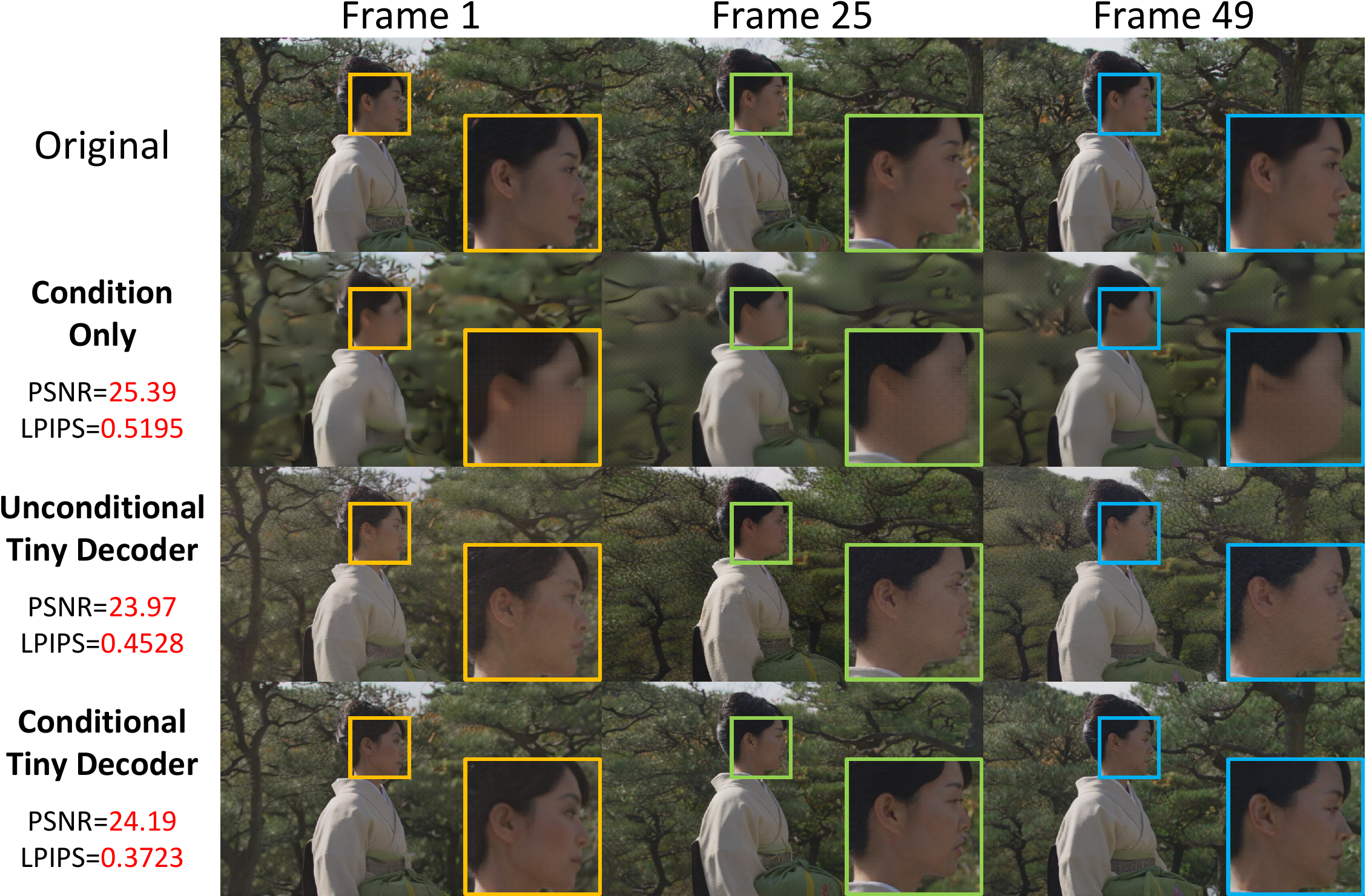}
    \caption{\textbf{Visual analysis on the codec condition and the conditional tiny decoder.} Videos are evaluated in 1080p and 0.0005 bpp.}
    \label{fig:cond}
\end{minipage}
\end{figure*}

\textbf{Preliminaries: Autoregressive Video Generation.} Autoregressive (AR) video diffusion predicts one latent frame (or chunk) at a time based on previous generation, involving three typical training paradigms. Teacher forcing (TF)~\citep{gao2024ca2} trains to denoise the next frame conditioned on a clean ground-truth prefix. It supports parallel training with a causal mask, but replaces that prefix with imperfect generated latents at inference, causing exposure bias. Diffusion forcing (DF)~\citep{chen2024diffusion} assigns different noise levels to frames in a training sequence, leaving a similar train-test gap~\citep{huang2025selfforcing,zhu2026causal}. Self forcing (SF)~\citep{huang2025selfforcing} closes this gap by performing an AR rollout during training and conditioning on previously generated clean latents, but serializes training and typically truncates gradients through the rollout. These paradigms were adopted primarily for video generation, where generated latents are the only strong temporal observations. It remains unclear which AR paradigm is most effective when a learned codec condition is jointly optimized.

\textbf{Autoregressive Paradigms with a Video Codec.} As shown in Fig.~\ref{fig:ar}, a codec permits two AR inference pipelines. \textit{AR with generated clean latents} follows conventional AR video generation and predicts $\hat l_t$ from the current codec condition $c_t$ and previously generated clean latents $\hat l_{<t}$. It corresponds to conditional TF or SF training. TF is parallel but train-test inconsistent, whereas SF is consistent but requires a sequential self-rollout, making joint optimization with a recurrent codec expensive. The second, \textit{AR with contextual latent}, is specific to video compression. Because the bitstream provides a frame-aligned condition at every time step, the model can instead use the causal condition sequence as history without feeding reconstructed latents back into later predictions. We train this path using a codec-conditioned variant of DF. Since $c_{\leq t}$ carries the temporal history rather than a differently noised latent prefix, all target latents can share a diffusion timestep and be optimized in parallel under causal attention. The conditions are decoded from the same bitstream during training and inference, so this formulation avoids conditioning on accumulated generation errors and allows every frame to provide gradients to the codec in parallel.

Contextual autoregression also follows established practice in neural video coding, where learned feature context is generally more effective for inter-frame prediction than a reconstructed frame alone~\citep{DCVC,DCVC-TCM,DCVC-DC}. An RGB frame has only three channels, and a clean latent in video VAE has only 16. By contrast, codec condition $c_t$ supports much more learned channels (e.g., 256). It is not constrained to be directly reconstructible and can preserve diverse predictive cues about motion, structure, and color. Guided by these insights, we adopt contextual autoregression trained with conditional DF for UniGVC-AR. Fig.~\ref{fig:ar_visual} validates this choice visually. At similar bitrates, DF better preserves appearance over time and exhibits less accumulated degradation than conditional TF and SF.

\textbf{Attention and Joint Objective.} Both diffusion branches are initialized from the Wan2.1-1.3B video DiT~\citep{wan2025wan}. BI-DiT preserves dense spatial-temporal attention. AR-DiT instead applies a causal temporal window of $w$ latent frames. This masked attention admits streaming inference and bounds the KV-cache length. We use $w=3$ based on the ablation in Sec.~\ref{sec:ablation}. LoRA adapters~\citep{hu2021lora} are inserted into the two DiTs to preserve the generative priors. We jointly update the adapters and codec using flow matching loss $\mathcal{L}_{\mathrm{FM}}$. The objective for stage 1 is
\begin{equation}
 \mathcal{L}_{\mathrm{Stage1}}=
 \lambda\cdot\mathcal{R}+\lambda_{\mathrm{FM}}\cdot\mathcal{L}_{\mathrm{FM}}.
 \label{eq:joint-objective}
\end{equation}
Gradients from both decoding modes consequently shape the same bitstream representation. After training, the dense branch supports sequence-level generation from a buffered coded segment, whereas the causal branch processes newly available $c_t$ with bounded KV memory.

\section{Toward Real-Time Autoregressive Video Diffusion Decoder}

\subsection{Adversarial Distribution Matching Distillation}
\label{sec:dmd}

The multi-step AR-DiT is still expensive for real-time. We distill it into one step using the stronger BI-DiT as the distribution teacher and adversarial supervisor under the same codec condition $c$. 

\textbf{Conditional Distribution Matching Distillation.} After merging the LoRAs, we freeze the codec and initialize the real-score network $\mathrm{s}_{\mathrm{real}}$ and the fake-score network $\mathrm{s}_{\mathrm{fake}}$ from BI-DiT, and optimize the full generator $\mathrm{G}_\theta$ and $\mathrm{s}_{\mathrm{fake}}$. For flow time $\tau\in[0,1]$ and noise $\epsilon\sim\mathcal{N}(0,I)$, we perturb the generated latent $\hat l$ into $\hat l_\tau=(1-\tau)\cdot\hat l+\tau\cdot\epsilon$. Following DMD~\citep{yin2024dmd,yin2024improved}, the gradient is estimated by the conditional score differences weighted by a normalizer $w(\tau)$:
\begin{equation}
 \nabla_\theta \mathcal{L}_{\mathrm{DMD}}
 =\mathbb{E}\!\left[
 w(\tau)\big(\mathrm{s}_{\mathrm{fake}}(\hat l_\tau,\tau;c)
 -\mathrm{s}_{\mathrm{real}}(\hat l_\tau,\tau;c)\big)
 \frac{\partial \mathrm{G}}{\partial\theta}
 \right].
 \label{eq:dmd}
\end{equation}

\textbf{Appending a Discriminator Head.} Distribution matching alone aligns the student with the imperfect teacher distribution. We therefore add a latent adversarial objective~\citep{lin2025diffusion} that directly distinguishes real and generated video latents. From intermediate features $F_i$ of $\mathrm{s}_{\mathrm{fake}}(\hat l_\tau,\tau;c)$, a discriminator head $\mathrm{D}_\phi$ with cross-attention, MLP, and learnable query $\mathrm{q}_i$ computes the scalar logit
\begin{equation}
 a_i=\mathrm{MLP}_i\!\left(\operatorname{Cross-Attn}(\mathrm{q}_i,\mathrm{k}_i(F_i),\mathrm{v}_i(F_i))\right),
 \qquad \mathrm{D}_\phi(\hat l_\tau,\tau;c)=\mathrm{MLP}\!\left([a_1;\ldots;a_n]\right).
 \label{eq:latent-discriminator}
\end{equation}
With the non-saturating GAN loss~\citep{goodfellow2014generative}, the adversarial and stage objectives are
\begin{align}
 \mathcal{L}_{\mathrm{GAN}}^{\mathrm{latent}} & =\mathbb{E}\!\left[\operatorname{softplus}(\mathrm{D}_\phi(l_\tau,\tau;c))\right] + \mathbb{E}\!\left[\operatorname{softplus}(-\mathrm{D}_\phi(\hat l_\tau,\tau;c))\right],
 \label{eq:latent-gan} \\
  \mathcal{L}_{\mathrm{Stage2}} & =\mathcal{L}_{\mathrm{DMD}}
 +\lambda_{\mathrm{GAN}}^{\mathrm{latent}}\cdot\mathcal{L}_{\mathrm{GAN}}^{\mathrm{latent}}.
\end{align}
During discriminator updates, adversarial gradients are stopped at the fake-score backbone, so the head $\mathrm{D}_\phi$ is optimized by classification while $\mathrm{s}_{\mathrm{fake}}$ is updated only by its flow matching loss. We perform five $\mathrm{D}_\phi$ and $\mathrm{s}_{\mathrm{fake}}$ updates per $\mathrm{G}_\theta$ update. These separations stabilize adversarial learning while retaining an accurate estimate of the generator's evolving distribution.

\subsection{Pixel Finetuning with Conditional Tiny Decoder}
\label{sec:tiny-decoder}

Once AR-DiT is reduced to one step, the original video VAE decoder becomes the dominant latency term because of its deep 3D causal convolutions and attention. We replace it in UniGVC-AR with a lightweight tiny decoder composed of 2D causal memory blocks and PixelShuffle upsampling~\citep{shi2016pixelshuffle,ultraflash2026}. Each block fuses its current-frame activation with a cached feature from the preceding frame, preserving causal temporal information without 3D convolution.

\textbf{Pixel Finetuning.} We optimize the entire tiny decoder together with newly inserted LoRA in the one-step AR-DiT. Given the reconstruction video $\hat x$. We construct a DINOv3~\citep{simeoni2025dinov3} feature discriminator to supply the pixel adversarial term $\mathcal{L}_{\mathrm{GAN}}^{\mathrm{pixel}}$, and improve temporal stability with flow warping error $\mathcal{L}_{\mathrm{warp}}$~\citep{lai2018learning} from a frozen RAFT~\citep{teed2020raft}. Perceptual supervision $\mathcal{L}_\mathrm{p}$ consists of LPIPS and DISTS measures from the VGG features~\citep{LPIPS,DISTS}. The final pixel stage objective is
\begin{align}
 \mathcal{L}_{\mathrm{Stage3}}&=\lambda_{\mathrm{MSE}}\cdot\|x-\hat{x}\|_2^2
 +\lambda_{\mathrm{p}}\cdot\mathcal{L}_{\mathrm{p}}(x,\hat{x})+\lambda_{\mathrm{GAN}}^{\mathrm{pixel}}\cdot\mathcal{L}_{\mathrm{GAN}}^{\mathrm{pixel}}(x,\hat{x})
 +\lambda_{\mathrm{warp}}\cdot\mathcal{L}_{\mathrm{warp}}(\hat{x}). \label{eq:pixel-loss}
\end{align}

\textbf{Conditional Tiny Decoder.} Inspired from the dual-branch decoding in image codec~\citep{zhang2025stablecodec,zhang2026ultra}, we make full use of the codec condition $c$ to enhance the tiny decoder, whose first projection now receives the concatenation $[\hat l_t;c_t]$. In Fig.~\ref{fig:cond}, we train the tiny decoder with condition only, generated latent only and their combination. The high-dimensional codec condition retains faithful geometry and color even at bitrates as low as 0.0005 bpp, and steadily improves the reconstruction quality when concatenated to condition the tiny decoder.

\section{Experiments}
\label{sec:experiments}

\subsection{Settings}

\textbf{Datasets.} Our training corpus contains 100K $512\times512$ images from MLIC-100K~\citep{jiang2023mlic}, 1M 720p clips sampled from Koala-36M~\citep{wang2025koala}, and 11K 1080p clips of at least 97 frames from UltraVideo~\citep{xue2025ultravideo}. We resize and center-crop samples to the resolution required by each training stage. Evaluation uses the first 96 frames of the 1080p sequences in HEVC-B~\citep{sullivan2012overview}, UVG~\citep{UVG}, and MCL-JCV~\citep{MCL-JCV}. Neural codecs receive RGB frames converted from YUV420 with BT.709.

\textbf{Implementation.} Codec initialization takes 100K steps on images, 300K steps on 29-frame clips, and 35K steps on 97-frame clips. The batch size is 8 and clip size is $256\times320$. Starting from $\lambda=0.25$, we finetune ultra-low-rate operating points with $\lambda\in\{1,2,4,8,16\}$. Stage 1 takes 15K steps on $256\times320$ clips and 2K steps on $512\times832$ with a batch size of 32. Each DiT is then adapted independently for 600 steps on $1024\times1920$ clips using only flow matching. Stage 2 runs for 600 steps on 97-frame clips, while stage 3 takes 2.6K steps on 25-frame clips, both using $1024\times1920$ clips and a batch size of 32. All models are trained in BF16 precision.

\textbf{Evaluation.} We compare with H.266/VVC using VTM-17.0 in low-delay (VTM-LD) and random-access (VTM-RA) configurations~\citep{bross2021overview}; distortion-oriented neural codecs DCVC-FM~\citep{DCVC-FM} and DCVC-RT~\citep{DCVC-RT}; and generative codecs GLC-Video~\citep{qi2025generative}, GLVC~\citep{glvc2025}, GNVC-VD~\citep{gnvcvd2025}, GVC1D~\citep{gvc1d2026}, Free-GVC~\citep{freegvc2026}, and S2VC~\citep{s2vc2025}. We measure distortion with PSNR and MS-SSIM~\citep{MS-SSIM}, perceptual fidelity with LPIPS~\citep{LPIPS}, DISTS~\citep{DISTS}, FID~\citep{heusel2017fid} and KID~\citep{binkowski2018demystifying}. Temporal behavior is measured by flow warping error $\mathrm{E}_{\mathrm{warp}}^*$ ($\times10^{-3}$)~\citep{lai2018learning}, and overall video quality by DOVER~\citep{wu2023exploring}. UniGVC-BI performs 50-step iterative denoising while UniGVC-AR applies 1-step denoising.

\subsection{Main Comparison}

\begin{figure*}[t]
    \centering
    \includegraphics[width=\textwidth]{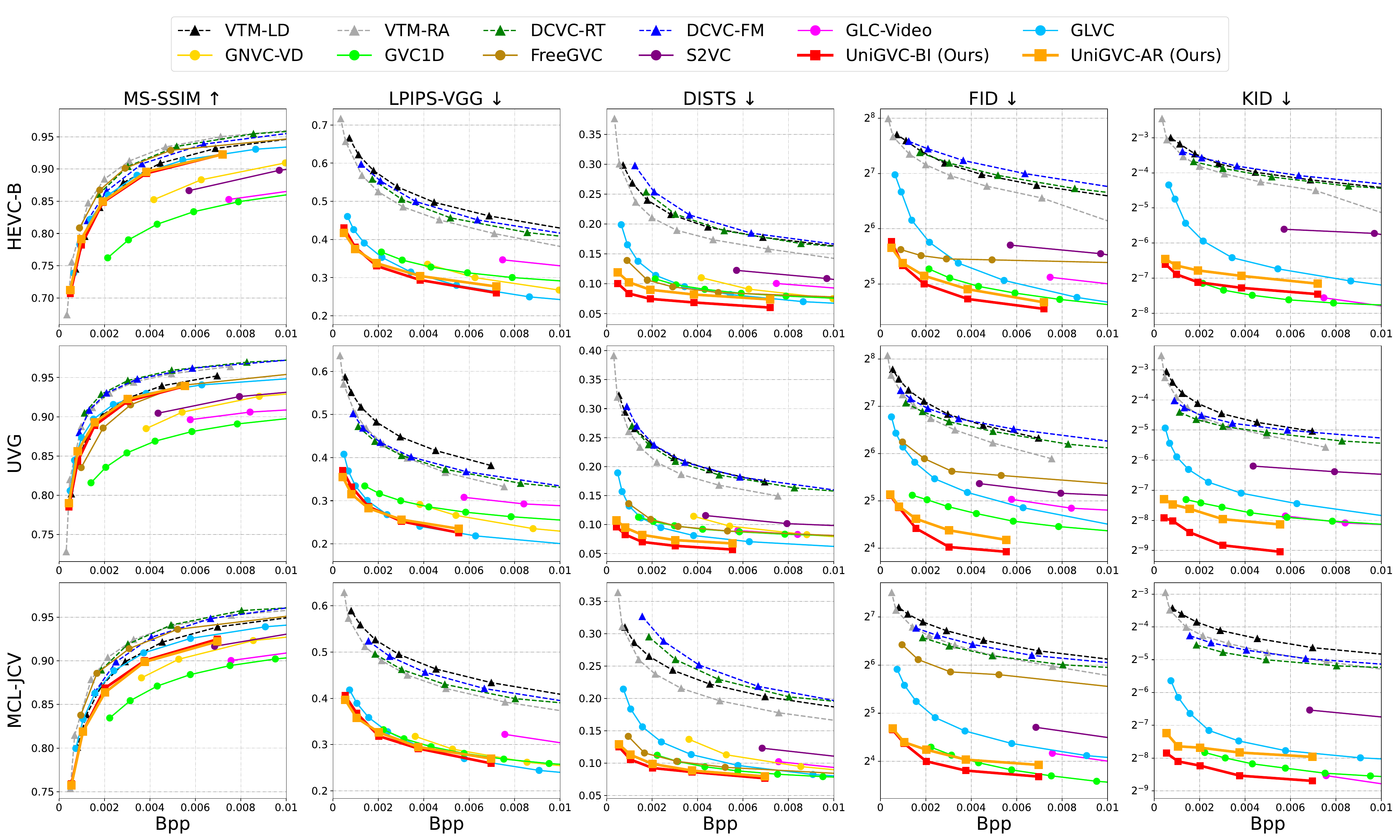}
    \caption{\textbf{Rate-distortion-perception curves on HEVC-B, UVG and MCL-JCV.} All video sequences are tested with the first 96 frames. UniGVC is evaluated with an intra-period of –1.}
    \label{fig:rdcurve}
\end{figure*}

\textbf{Rate-distortion-perception performance.} Fig.~\ref{fig:rdcurve} compares all methods over the ultra-low-rate region. UniGVC-BI forms the strongest overall frontier on all three datasets. It preserves structures while obtaining markedly better perceptual fidelity than MSE codecs at comparable rates, and it also improves the distributional FID and KID criteria over generative codecs. The gain is especially pronounced near the lowest operating point of 0.0005 bpp. UniGVC-AR deliberately trades some of the bidirectional model's quality for latency, yet remains a competitive overall performance.

\begin{figure*}[t]
\centering
\begin{minipage}[t]{0.491\textwidth}
    \centering
    \includegraphics[width=\linewidth]{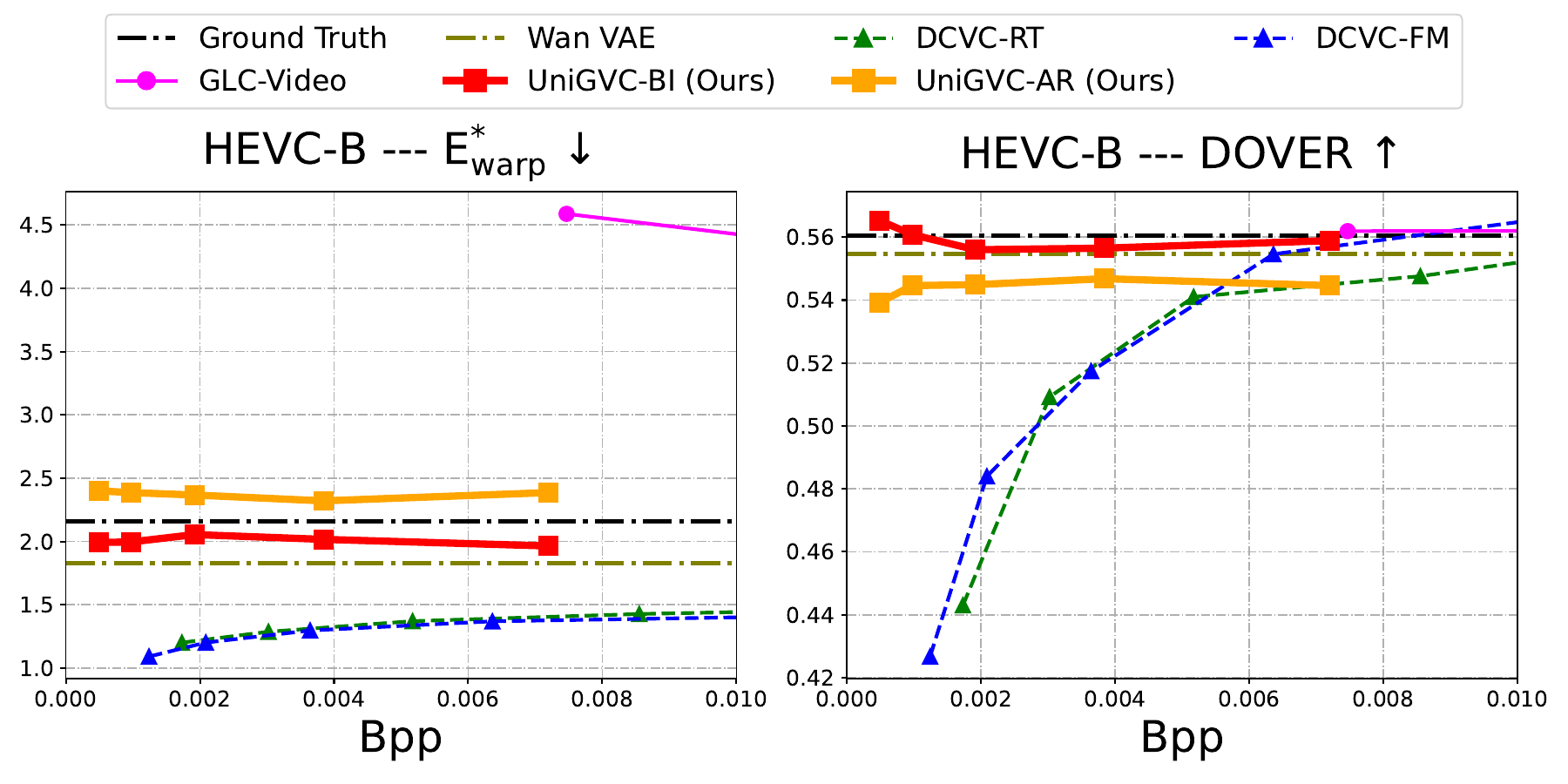}
    \caption{Temporal stability and video quality.}
    \label{fig:rdcurve_temporal}
\end{minipage}\hfill
\begin{minipage}[t]{0.49\textwidth}
    \centering
    \includegraphics[width=\linewidth]{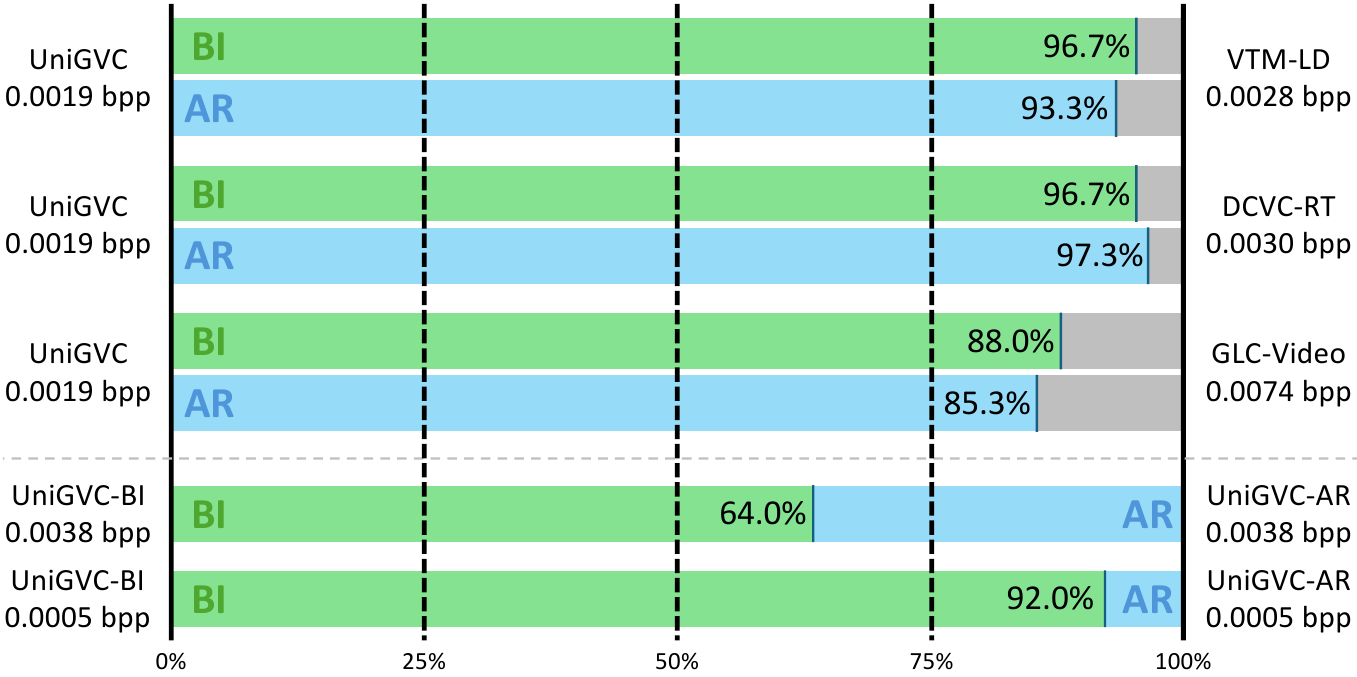}
    \caption{Human preference study on HEVC-B}
    \label{fig:human}
\end{minipage}
\end{figure*}

\textbf{Temporal quality.} As shown in Fig.~\ref{fig:rdcurve_temporal}, GLC-Video's frame-wise reconstruction incurs substantially larger warping error, illustrating that sharp individual frames do not guarantee a stable video. UniGVC-BI remains much closer to the temporally conservative distortion codecs while delivering higher DOVER quality. More importantly, its curve changes smoothly as the rate falls, including at 0.0005 bpp, rather than exhibiting a low-rate flicker collapse. UniGVC-AR is less stable than the bidirectional mode, but retains a competitive balance between temporal consistency and DOVER.

\begin{figure*}[t]
    \centering
    \includegraphics[width=\textwidth]{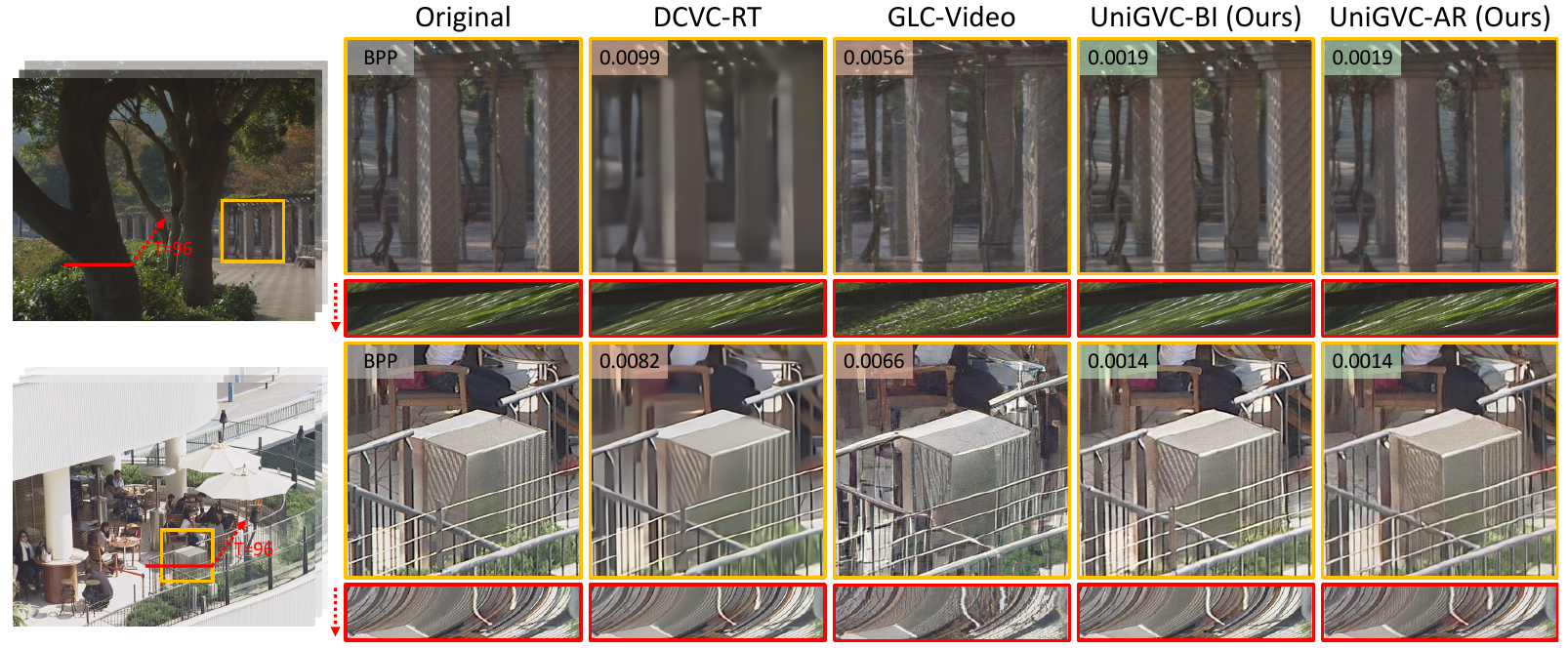}
    \caption{\textbf{Qualitative comparison on HEVC-B.} Each case displays the temporal profile (red box) at one row across 96 frames to observe temporal coherence. \textit{Best viewed on screen for details.}}
    \label{fig:visual}
\end{figure*}

\textbf{User Study.} Fig.~\ref{fig:human} records human preference of reconstructed HEVC-B videos under ultra-low bitrates. The study presents the ground truth video, and two reconstructions from different methods in random orders on the same high-resolution screen, collecting 30 votes from human observers for each comparison. Users express a clear preference for UniGVC over VTM-LD, DCVC-RT and GLC-Video, and an obvious tendency on UniGVC-BI against UniGVC-AR at lower bitrates. 

\textbf{Visual Comparison.} Fig.~\ref{fig:visual} presents the quantitative results. Despite using substantially lower rates, UniGVC-BI recovers temporally stable texture and clean boundaries. UniGVC-AR preserves the same scene layout and motion with slight temporal artifacts, probably resulting from the tiny decoder where temporal growth adopts 2D convolutions. In contrast, DCVC-RT remains temporally steady but smooths thin edges and repeated texture, whereas GLC-Video synthesizes sharper content whose fine structures vary across frames and appear discontinuous in the temporal strips.

\subsection{Ablation Study}
\label{sec:ablation}

\begin{wraptable}{r}{0.48\textwidth}
\caption{\textbf{Ablation on DiT attention.} BD-rate is measured on HEVC-C with 50-step denoising. FPS is measured on 1080p videos with 1 step.}
\centering
\setlength{\tabcolsep}{3pt}
\renewcommand{\arraystretch}{1.25}
\footnotesize
\begin{tabular}{c|c|cc}
\Xhline{1.0pt}
\multirow{2}{*}{Attention} & \multirow{2}{*}{DiT FPS} & \multicolumn{2}{c}{BD-rate ($\downarrow\%$)} \\
\cline{3-4}
~ & ~ & LPIPS & DISTS \\
\Xhline{1.0pt}
Bidirectional & 7.09 & \textbf{-29.13} & \textbf{-60.71} \\
\hline
Causal & 24.45 & 0 & 0 \\
\hline
Causal ($w=5$) & 44.92 & +2.63 & +2.57 \\
\textbf{Causal ($\bm{w=3}$)} & 62.18 & +4.96 & +5.35 \\
Causal ($w=2$) & 66.37 & +12.25 & +17.33 \\
Causal ($w=1$) & \textbf{68.23} & +40.11 & +42.63 \\
\Xhline{1.0pt}
\end{tabular}
\label{tab:ablation:attention}
\end{wraptable}

\paragraph{Bidirectional and causal attention.}
Table~\ref{tab:ablation:attention} examines how attention types affect the end-to-end performance. We train codec with the DiT using different attention. Training objectives include the rate loss and the flow matching loss. We evaluate BD-rate~\citep{bd-rate} with four operating points in 0.001-0.004 bpp. Dense bidirectional attention offers the best compression efficiency, but runs at only 7.09 FPS. Causal attention reaches 24.45 FPS, and bounding its history yields a further large speedup at the cost of a consistently performance degradation. A window size of three reaches 62.18 FPS. Further reduction saves little additional time but sharply degrades quality. We therefore set $w=3$ for UniGVC-AR to strike a balance between speed and quality.

\newcolumntype{"}{@{\hskip\tabcolsep\vrule width
1pt\hskip\tabcolsep}}
\begin{table*}[t]
\caption{\textbf{Ablation on decoding acceleration and enhancement.} We examine how architectural designs affect decoding speed, including 50-step bidirectional DiT (\textbf{BI-50}), 1-step autoregressive DiT (\textbf{AR-1}), original Wan VAE decoder (\textbf{Orig.}), and tiny decoder (\textbf{Tiny}). We also study how training strategies improve the perceptual fidelity, including adversarial distribution matching distillation (\textbf{Adv. DMD}), pixel finetuning (\textbf{Pixel FT.}), and conditional tiny decoder (\textbf{Cond. VAE-D}). Decoding speed and quality are evaluated on HEVC-B at 1080p, using four bitrates in 0.0005-0.004 bpp. Final choises for \colorbox{green!15}{UniGVC-BI} and \colorbox{cyan!15}{UniGVC-AR} are highlighted in color.}
\centering
\setlength{\tabcolsep}{3pt}
\renewcommand{\arraystretch}{1.25}
\footnotesize
\begin{tabular}{c|cc|ccc"ccc|c|cc}
\Xhline{1.0pt}
\multirow{3}{*}{Exp.} & \multicolumn{2}{c|}{Architecture} & \multicolumn{3}{c"}{Training Strategy} & \multicolumn{3}{c|}{Latency (ms / frame)} & \multirow{3}{*}{FPS} & \multicolumn{2}{c}{Bj{\o}ntegaard Delta} \\
\cline{2-9}
\cline{11-12}
~ & \multirow{2}{*}{DiT} & \multirow{2}{*}{VAE-D} & Adv. & Pixel & Cond. & \multirow{2}{*}{Codec} & \multirow{2}{*}{DiT} & \multirow{2}{*}{VAE-D} & ~ & \multirow{2}{*}{BD-LPIPS$\downarrow$} & \multirow{2}{*}{BD-DISTS$\downarrow$} \\
~ & ~   & ~     & DMD  & FT.   & VAE-D & ~     & ~   & ~     & ~ & ~ & ~ \\
\Xhline{1.0pt}
\rowcolor{green!15}
1 & BI-50 & Orig. & - & - & - & 7.42 & 6948.61 & 69.34 & 0.14 & \textbf{0} & \textbf{0} \\
\hline
2 & AR-1  & Orig. & - & - & - & 7.42 & 16.10 & 69.34 & 10.77 & +0.069 & +0.142 \\
3 & AR-1  & Orig. & \checkmark & - & - & 7.42 & 16.10 & 69.34 & 10.77 & +0.041 & +0.045 \\
\hline
4 & AR-1  & Tiny  & \checkmark & - & - & 7.42 & 16.10 & 3.62 & 36.85 & +0.074 & +0.068 \\
5 & AR-1  & Tiny  & \checkmark & \checkmark & - & 7.42 & 16.10 & 3.62 & 36.85 & +0.058 & +0.048 \\
6 & AR-1  & Tiny  & - & \checkmark & \checkmark & 7.42 & 16.10 & 3.62 & 36.85 & +0.075 & +0.079 \\
\rowcolor{cyan!15}
7 & AR-1  & Tiny  & \checkmark & \checkmark & \checkmark & \textbf{7.42} & \textbf{16.10} & \textbf{3.62} & \textbf{36.85} & +0.001 & +0.017  \\
\Xhline{1.0pt}
\end{tabular}
\label{tab:ablation:decoder}
\end{table*}

\paragraph{From slow bidirectional to fast autoregressive decoding.}
Table~\ref{tab:ablation:decoder} tracks the complete decoding acceleration and enhancement. Replacing BI-50 by AR-1 reduces DiT latency from 6948.61 to 16.10 ms per frame, but the direct one-step inference leaves substantial quality degradation. Adversarial DMD narrows these gaps, confirming that a bidirectional score teacher can improve a causal student. At that point the original VAE decoder takes 69.34 ms and dominates the pipeline. The tiny decoder reduces this component to 3.62 ms and raises end-to-end decoding throughput from 10.77 to 36.85 FPS, although a direct decoder replacement largely hurts quality. Pixel fine-tuning and codec conditioning recover most of this loss: the final UniGVC-AR lags behind UniGVC-BI by only 0.001 in BD-LPIPS and 0.017 in BD-DISTS on average. However, these small metric gaps between the two modes do not necessarily reflect the actual difference in perceived visual quality.

\section{Related Work}

\subsection{Generative Video Compression}

Early generative codecs introduced adversarial losses into image and video compression~\citep{agustsson2019generative,mentzer2020high,yang2022plvc}. Inspired by image codecs~\citep{jia2024generative,xue2026one,zhang2025stablecodec,zhang2026ultra}, recent video methods witnessed a growing trend on exploiting pretrained generative latent spaces or diffusion priors. GLC-Video compresses discrete generative latents~\citep{qi2025generative}, whereas GLVC uses a continuous video tokenizer and recurrent latent coding~\citep{glvc2025}. DiffVC applies image-diffusion for perceptual enhancement~\citep{ma2025diffusion}. Concurrent designs explore compact one-dimensional latents~\citep{gvc1d2026} or a single-step image diffusion decoder~\citep{s2vc2025}. Video diffusion models display strong potential in alleviating temporal flickering. To this end, GNVC-VD performs sequence-level latent refinement with a video DiT~\citep{gnvcvd2025}; and Free-GVC codes trajectories of a pretrained video diffusion model without task-specific training~\citep{freegvc2026}. These approaches advance perceptual quality, but expose only one fixed decoding trade-off. UniGVC instead transmits a decoder-agnostic condition that can drive either a multi-step bidirectional generator or a one-step causal generator.

\subsection{Autoregressive Video Diffusion and Acceleration}

Standard video diffusions use dense bidirectional attention and iterative sampling, which provide strong sequence-level coherence but prevent online decoding. CausVid converts a bidirectional DiT into a few-step causal generator and applies asymmetric distribution-matching distillation (DMD)~\citep{yin2025causvid,yin2024dmd,yin2024improved}. Self Forcing addresses exposure bias by training on autoregressively generated histories~\citep{huang2025selfforcing}, while Causal Forcing studies the architectural mismatch between bidirectional teachers and causal students~\citep{zhu2026causal}. Adversarial post-training provides another route to one-step video synthesis~\citep{lin2025diffusion}. These advances target open-ended generation, where past generated frames are the principal temporal condition and training must carefully reproduce their inference-time distribution.

\section{Conclusion}

This work presented UniGVC, a unified generative video compression framework that decouples the transmitted representation from the receiver's quality-latency choice for ultra-low-rate scenarios. A hierarchical latent codec produces one bitstream and one sequence of compression-aware conditions, driving either a multi-step bidirectional DiT for high-quality random-access decoding or a distilled one-step autoregressive DiT with a causal tiny decoder for low-delay streaming. Context-based autoregression enables parallel, train-test-consistent optimization, while adversarial distribution matching and pixel fine-tuning preserve much of the quality after acceleration. Across 1080p benchmarks, UniGVC-BI provides the strongest overall rate-distortion-perception trade-off among the evaluated methods, and UniGVC-AR reaches 36.85 FPS with only a small perceptual gap.

\textbf{Limitations.}
Although UniGVC avoids storing or transmitting duplicate bitstreams, the retaining two decoding parameter sets increases the storage of decoder-side model. In addition, as with generative codecs, plausible detail may not reproduce the source exactly at extreme rates. Applications requiring scientific fidelity should avoid aggressive operating points or transmit additional residual information. Future work includes shared-weight multi-mode DiTs and region enhancement.

\clearpage

\bibliography{iclr2027_conference}
\bibliographystyle{iclr2027_conference}

\clearpage
\appendix

\section{Test Settings}
\label{app:test-settings}

\textbf{Test data.}
The source sequences of HEVC-B~\citep{sullivan2012overview}, UVG~\citep{UVG}, and MCL-JCV~\citep{MCL-JCV} are stored in YUV~4:2:0 format. We use the first 96 frames of every sequence. Traditional codecs operate directly on the original YUV data. For neural codecs, we convert the source frames to RGB using the BT.709 conversion matrix before encoding. The decoded outputs of traditional codecs are converted to RGB with the same procedure, and all objective metrics are computed between RGB reference and reconstruction frames. 

\textbf{Traditional codec.}
We evaluate H.266/VVC~\citep{bross2021overview} with the VTM-17.0 reference software. VTM-LD uses the Low-Delay-B configuration \texttt{encoder\_lowdelay\_vtm.cfg} with
$\mathrm{QP}\in\{37,40,43,46,49,52,55\}$. VTM-RA uses \texttt{encoder\_randomaccess\_vtm.cfg} with $\mathrm{QP}\in\{36,39,42,45,48,51,57,63\}$. Encoding and decoding are performed in the native YUV~4:2:0 format; only the decoded sequence used for metric computation is converted to RGB.

\textbf{Neural codecs.}
For the distortion-oriented baselines DCVC-FM~\citep{DCVC-FM} and DCVC-RT~\citep{DCVC-RT}, and for GLC-Video~\citep{qi2025generative}, we run the official code and released checkpoints locally. All frames are processed in RGB. The code or checkpoints of GLVC~\citep{glvc2025}, GNVC-VD~\citep{gnvcvd2025}, GVC1D~\citep{gvc1d2026}, Free-GVC~\citep{freegvc2026}, and S2VC~\citep{s2vc2025} were not publicly available when our evaluation was completed. We therefore use rate-distortion-perception points provided directly by the authors for GLVC, GVC1D, and S2VC, and the available results reported in the papers for GNVC-VD and Free-GVC.

\textbf{Metrics.}
We evaluate PSNR, MS-SSIM~\citep{MS-SSIM}, and DISTS~\citep{DISTS} with PyIQA under its default settings. LPIPS~\citep{LPIPS}, FID~\citep{heusel2017fid} and KID~\citep{binkowski2018demystifying} are implemented with TorchMetrics. Since benchmarks contain relatively few high-resolution frames, FID and KID are computed from overlapping $256\times256$ RGB patches extracted from the original and reconstructed sequences. We use the official implementation for the flow warping error $\mathrm{E}_{\mathrm{warp}}^*$~\citep{lai2018learning} and DOVER~\citep{wu2023exploring}. Lower LPIPS, DISTS, FID, KID, and $\mathrm{E}_{\mathrm{warp}}^*$ are better, whereas higher PSNR, MS-SSIM, and DOVER are better.

\section{Implementation Details of UniGVC}
\label{app:implementation}

\subsection{Codec Design}
\label{app:codec-design}

Table~\ref{tab:codec-architecture} summarizes the primary codec modules. Spatial resolutions are measured relative to the input video: the Wan VAE first maps a frame to $1/8$ spatial resolution, the primary latent $y_t$ is represented at $1/16$, and the hierarchical hyperlatents $(r_t,z_t)$ are represented at $1/64$. The primary latent contains 256 channels. The hyper-analysis output contains 128 channels and is split into a 16-channel residual branch $r_t$ and a 112-channel context-modeled branch $z_t$.

The codec is built mainly from depthwise convolution (DC) blocks~\citep{dcvcuf2026}. Each block contains a residual $3\times3$ depthwise convolution followed by two $1\times1$ pointwise projections and a weighted SiLU activation; a second residual feed-forward branch expands the feature width by four. Downsampling uses PixelUnshuffle followed by a $1\times1$ projection, while upsampling uses a $1\times1$ projection followed by PixelShuffle. The latent analysis and synthesis transforms share their main parameters between intra and inter latent frames. Only entropy parameters are specialized for the first intra latent frame, and the rest inter latent frames.

\begin{table*}[h]
\caption{\textbf{Architecture of the primary modules in the hierarchical latent video codec.} ``Dimension'' denotes the dominant internal feature width, and spatial resolution is relative to the input.}
\label{tab:codec-architecture}
\centering
\setlength{\tabcolsep}{3pt}
\renewcommand{\arraystretch}{1.22}
\footnotesize
\begin{tabular}{lccccccccc}
\toprule
 Module & \makecell{Latent\\Encoder} & \makecell{Latent\\Decoder} & \makecell{Hyper\\Encoder} & \makecell{Hyper\\Decoder} & \makecell{$y$ Entropy\\Model} & \makecell{$z$ Entropy\\Model} & \makecell{Context\\Extractor} & \makecell{Memory\\Fuser} & \makecell{Condition\\Head} \\
\midrule
Dimension & 256 & 256 & 128 & 128 & 512 & 256 & 512, 256 & 512 & 256 \\
DC blocks & 8 & 8 & 2 & 2 & 4 & 4 & 4, 4 & 4 & 4 \\
Resolution & $\frac18\!\rightarrow\!\frac1{16}$ & $\frac1{16}\!\rightarrow\!\frac18$ & $\frac1{16}\!\rightarrow\!\frac1{64}$ & $\frac1{64}\!\rightarrow\!\frac1{16}$ & $\frac1{16}$ & $\frac1{64}$ & $\frac18$ & $\frac18$ & $\frac18$ \\
\bottomrule
\end{tabular}
\end{table*}

\subsection{Training Procedure}
\label{app:training}

Table~\ref{tab:training-schedule} details the complete training curriculum. All resources are rented individually.

\textbf{Codec initialization.}
We first train the codec independently of the diffusion models. The curriculum uses $(1,29,97,97,97)$ frames for $(100\mathrm{K},300\mathrm{K},20\mathrm{K},10\mathrm{K},5\mathrm{K})$ steps, with learning rates $(10^{-4},10^{-4},5\!\times\!10^{-5},2\!\times\!10^{-5},10^{-5})$. The image stage uses $512\times512$ crops; all video stages use $256\times320$ crops and batch size 8. We first obtain an initialization with $\lambda=0.25$ in Eq.~\ref{eq:codec-pretrain}, then derive the ultra-low-rate models using $\lambda\in\{1,2,4,8,16\}$ and additional 97-frame training. The hierarchical weights~\citep{glvc2025} for the first four VAE latent frames are $(12.0,1.2,0.5,0.9)$; subsequent frames alternate between 0.5 and 0.9.

\textbf{Stage 1: joint codec--diffusion training.}
Each rate point resumes from its codec checkpoint. We insert rank-32 LoRA modules into BI-DiT and AR-DiT. For the first 15K steps on 97-frame $256\times320$ clips, the codec is frozen and only the LoRAs are optimized. We then jointly optimize the codec and both LoRAs for 5K steps at $256\times320$ and 2K steps at $512\times832$, using batch size 32. Independent LoRA adaptation uses learning rate $10^{-4}$, while joint optimization uses $10^{-5}$. We set $\lambda_{\mathrm{FM}}=0.35$, the flow-time shift to 5.0, and classifier-free condition dropout to 0.1. The rate weight of the first latent frame is multiplied by 0.1 during joint training to preserve a stronger causal anchor. Finally, with the codec fixed, we adapt the two LoRAs independently for 600 steps on 97-frame $1024\times1920$ clips using only flow matching. UniGVC-BI is fixed after this stage.

\textbf{Stage 2: adversarial distribution matching.}
We merge the stage-1 AR-DiT LoRA into its backbone and optimize the full one-step generator, online fake-score model, and discriminator head for 600 steps using 97-frame $1024\times1920$ clips and batch size 32. The codec and real-score BI-DiT remain frozen. We use learning rate $10^{-5}$ and $\lambda_{\mathrm{GAN}}^{\mathrm{latent}}=0.05$. The discriminator pools fake-score features after one-third, two-thirds, and all of the DiT depth. We update the discriminator head and fake-score model five times for every generator update; adversarial gradients are stopped at the fake-score backbone during discriminator optimization, while the fake-score backbone itself is trained with standard flow matching to model the distribution of the one-step generator.

\textbf{Stage 3: pixel fine-tuning.}
Pixel training uses 25-frame $1024\times1920$ clips, batch size 32, and 2.6K steps. The four phases contain $(1000,1000,500,100)$ steps with learning rates $(10^{-4},5\!\times\!10^{-5},2\!\times\!10^{-5},10^{-5})$. We train the entire tiny decoder and new rank-8 LoRA modules inserted into the one-step AR-DiT, while keeping the codec fixed. The DINOv3 discriminator input is resized to 448 to control memory consumption and follows the discriminator design of~\citet{zhang2026ultra}. In Eq.~\ref{eq:pixel-loss}, $(\lambda_{\mathrm{MSE}},\lambda_{\mathrm{p}},\lambda_{\mathrm{GAN}}^{\mathrm{pixel}},\lambda_{\mathrm{warp}})=(0.5,2.0,0.1,1.0)$; LPIPS-VGG and DISTS contribute equally within the perceptual loss $\mathcal{L}_{\mathrm{p}}$.

\begin{table*}[t]
\caption{Stage-wise training schedule.}
\label{tab:training-schedule}
\centering
\setlength{\tabcolsep}{2.5pt}
\footnotesize
\begin{tabular}{llcccc}
\toprule
Phase & Trainable modules & Frames & Crop size & Steps & Learning rate \\
\midrule
Codec initialization & Codec & 1 & $512\times512$ & 100K & $10^{-4}$ \\
Codec initialization & Codec & 29 & $256\times320$ & 300K & $10^{-4}$ \\
Codec initialization & Codec & 97 & $256\times320$ & \makecell{20K,\\10K,5K} & \makecell{$5\!\times\!10^{-5}$ ,\\$2\!\times\!10^{-5}$,$10^{-5}$} \\
\midrule
Stage 1 warm-up & BI/AR-DiT LoRA & 97 & $256\times320$ & 15K & $10^{-4}$ \\
Stage 1 joint & Codec + BI/AR-DiT LoRA & 97 & $256\times320$ & 5K & $10^{-5}$ \\
Stage 1 joint & Codec + BI/AR-DiT LoRA & 97 & $512\times832$ & 2K & $10^{-5}$ \\
Stage 1 adaptation & BI-DiT LoRA or AR-DiT LoRA & 97 & $1024\times1920$ & 600 each & $10^{-4}$ \\
\midrule
Stage 2 & AR-DiT + Fake score + GAN & 97 & $1024\times1920$ & 600 & $10^{-5}$ \\
Stage 3 & AR-DiT LoRA + Tiny decoder + GAN & 25 & $1024\times1920$ & \makecell{1K,1K,\\500,100} & \makecell{$10^{-4}$,$5\!\times\!10^{-5}$ ,\\$2\!\times\!10^{-5}$,$10^{-5}$} \\
\bottomrule
\end{tabular}
\end{table*}

\subsection{Inference with Different Decoding Modes}
\label{app:inference}

Both modes use exactly the same encoder, entropy-coded representation, textual side information, and recurrent codec decoder. After entropy decoding reconstructs the condition sequence, the receiver selects a generative path without transcoding the video bitstream. For UniGVC-BI, the conditions of the requested coding unit are buffered and processed jointly by BI-DiT with dense bidirectional attention. We perform 50 denoising steps with classifier-free guidance (CFG)~\citep{ho2022classifier} scale 3.0, then map the generated latent sequence to RGB with the original Wan video VAE decoder. This mode targets quality-oriented random-access playback. UniGVC-AR instead applies a single causal AR-DiT evaluation per latent frame without CFG. It maintains a windowed KV cache containing the most recent latent frames and immediately decodes each generated latent with the conditional tiny decoder. The bounded cache makes DiT computation linear in sequence length. Because the Wan latent space uses $4\times$ temporal downsampling and the tiny decoder performs the corresponding causal upsampling, streaming has an intrinsic four-frame output latency.

\section{Complexity Analysis}
\label{app:complexity}

\subsection{Module Parameters}
\label{app:parameters}

Table~\ref{tab:module-parameters} compares module parameter counts with generative codecs that use the same Wan VAE or video-diffusion family. Wan2.1-1.3B~\citep{wan2025wan} is included as a reference. GNVC-VD's DiT count includes its added control modules. The frozen text encoder is excluded from every row. We discuss caption generation and transmission separately in Sec.~\ref{app:text-prompt}. The tiny decoder in UniGVC-AR reduces pixel-decoder parameters from 72.6M to 13.5M, a $5.38\times$ reduction. The codec is shared between UniGVC-BI and UniGVC-AR, whereas their DiTs are separate mode-specific checkpoints. Thus, supporting both modes does not duplicate the transmitted representation, although a receiver that installs both modes must store both DiT parameter sets.

\begin{table}[t]
\caption{\textbf{Module parameter counts in millions.} We compare UniGVC with generative codecs GLVC and GNVC-VD implemented on a similar Wan2.1-1.3B framework. GLVC and GNVC-VD use reported module parameters from their papers due to unavailable code and checkpoints.}
\label{tab:module-parameters}
\centering
\footnotesize
\begin{tabular}{lccccc}
\toprule
Method & VAE Encoder & Codec & DiT modules & VAE Decoder & Total \\
\midrule
Wan2.1-1.3B~\citep{wan2025wan} & 53.6 & - & 1419.0 & 72.6 & 1545.2 \\
GLVC~\citep{glvc2025} & 53.6 & 100.4 & - & 72.6 & 226.6 \\
GNVC-VD~\citep{gnvcvd2025} & 53.6 & 53.1 & 2154.5 & 72.6 & 2333.8 \\
\midrule
UniGVC-BI & \multirow{2}{*}{\makecell{53.6\\(Shared)}} & \multirow{2}{*}{\makecell{82.0\\(Shared)}} & 1419.0 & 72.6 & 1627.2 \\
UniGVC-AR & ~ & ~ & 1419.0 & 13.5 & 1568.1 \\
\bottomrule
\end{tabular}
\end{table}

\subsection{Practical Latency}
\label{app:latency}

Table~\ref{tab:latency} reports average conding speed at different resolutions. Encoding is shared by the two modes and includes the Wan VAE encoder, latent codec, and entropy encoding. Decoding includes entropy decoding, generative latent reconstruction, and pixel decoding. At $1920\times1080$, replacing 50-step BI-DiT and the original VAE decoder with one-step AR-DiT and the tiny decoder reduces latency from approximately 7025 to 27 ms per frame, a $250\times$ speedup. The unrounded measurement corresponds to the 36.85 FPS reported in the main paper. UniGVC-AR remains real-time at all tested resolutions, while UniGVC-BI deliberately spends substantially more computation for its quality-oriented operating point. The current 1080p encoder takes 86 ms per frame.

\begin{table}[t]
\caption{Encoding and decoding speed per frame at different resolutions.}
\label{tab:latency}
\centering
\footnotesize
\begin{tabular}{ccccc}
\toprule
 & Decoding mode & $1920\times1080$ & $1080\times720$ & $640\times480$ \\
\midrule
Encoding & - & 86 ms & 39 ms & 21 ms \\
\midrule
Decoding & UniGVC-BI & 7025 ms & 1502 ms & 410 ms \\
Decoding & UniGVC-AR & 27 ms & 19 ms & 12 ms \\
\bottomrule
\end{tabular}
\end{table}

\section{Textual Prompt}
\label{app:text-prompt}

\subsection{Video Caption}
\label{app:caption}

The pretrained Wan DiT is text-conditioned. At the encoder, we therefore use Qwen3.5-397B-A17B to generate one sequence-specific caption for each 96-frame test video. This caption is transmitted as a losslessly compressed UTF-8 string and recovered exactly at the decoder. Both UniGVC modes concatenate it with the same fixed general quality prompt. The frozen Wan T5 encoder then maps the complete prompt to at most 512 text tokens; only the sequence-specific text, rather than its embedding tensor, is signaled in the bitstream.

We use the following captioning instruction:
\begin{quote}
\small
``Watch the video carefully and write one detailed, factual paragraph of 80--150 words. When applicable, describe (i) the setting and location, (ii) the main subjects and their appearance, (iii) actions, motion, and event progression, (iv) camera movement and framing, and (v) lighting, color tone, and atmosphere. Do not add a preamble, bullet points, headings, or quotation marks. Be objective, specific, and dense in detail, and avoid vague filler.''
\end{quote}
This instruction encourages the sequence caption to describe information useful for conditional reconstruction rather than subjective interpretation. We use the following 24-word general quality prompt for every sequence:
\begin{quote}
\small
``Photorealistic video, realistic motion, natural environmental dynamics, physically plausible interactions, consistent temporal behavior, cinematic composition, documentary-style realism, high visual fidelity, fine-grained details, 4K quality.''
\end{quote}
Because this general prompt is fixed and shared by all videos, it is stored as part of the codec configuration and incurs no per-sequence rate. Table~\ref{tab:caption-statistics} summarizes the sequence-specific captions used in our evaluation. They contain 4.7-5.3 sentences and 84.8-124.3 words on average. After concatenating the fixed prompt, the complete prompts contain 108.8, 136.4, and 148.3 words on HEVC-B, UVG, and MCL-JCV, respectively. The transmitted sequence-caption strings average 517.8-759.6 bytes before lossless compression.

\begin{table}[t]
\caption{\textbf{Statistics of the sequence-specific captions and complete prompts.} Averages are computed per sequence. The complete prompt includes the fixed 24-word general prompt, whereas the original byte count excludes it and corresponds to the original size in Table~\ref{tab:prompt-compression}.}
\label{tab:caption-statistics}
\centering
\setlength{\tabcolsep}{3.2pt}
\renewcommand{\arraystretch}{1.16}
\footnotesize
\begin{tabular}{cccccc}
\toprule
Dataset & \makecell{No. of\\sequences} & \makecell{Sentences\\per caption} & \makecell{Sequence\\words} & \makecell{Full-prompt\\words} & \makecell{Original\\bytes} \\
\midrule
HEVC-B & 5 & 4.8 & 84.8 & 108.8 & 517.8 \\
UVG & 7 & 4.7 & 112.4 & 136.4 & 684.9 \\
MCL-JCV & 30 & 5.3 & 124.3 & 148.3 & 759.6 \\
\bottomrule
\end{tabular}
\end{table}

\begin{table*}[t]
\caption{\textbf{Lossless compression of the transmitted sequence-specific captions.} Ratios are relative to the original UTF-8 string. Extra bpp amortizes one caption over 96 video frames at $1920\times1080$.}
\label{tab:prompt-compression}
\centering
\setlength{\tabcolsep}{5pt}
\renewcommand{\arraystretch}{1.16}
\footnotesize
\resizebox{\textwidth}{!}{%
\begin{tabular}{ccccccccc}
\toprule
Dataset & \makecell{Original\\bytes} & \makecell{Zlib\\bytes} & \makecell{Zstd+dict.\\bytes} & \makecell{Zlib\\ratio} & \makecell{Zstd\\ratio} & \makecell{Original\\extra bpp} & \makecell{Zlib\\extra bpp} & \makecell{Zstd+dict.\\extra bpp} \\
\midrule
HEVC-B & 517.8 & 305.0 & 184.2 & 0.5890 & 0.3557 & $2.081\!\times\!10^{-5}$ & $1.226\!\times\!10^{-5}$ & $7.403\!\times\!10^{-6}$ \\
UVG & 684.9 & 391.6 & 233.4 & 0.5718 & 0.3408 & $2.752\!\times\!10^{-5}$ & $1.574\!\times\!10^{-5}$ & $9.380\!\times\!10^{-6}$ \\
MCL-JCV & 759.6 & 426.5 & 245.2 & 0.5614 & 0.3228 & $3.053\!\times\!10^{-5}$ & $1.714\!\times\!10^{-5}$ & $9.854\!\times\!10^{-6}$ \\
\bottomrule
\end{tabular}}
\end{table*}

\subsection{Prompt Compression}
\label{app:prompt-compression}

We train a 512-KiB zstandard dictionary from one million non-empty captions sampled from the Koala-36M training corpus. The dictionary is fixed and distributed once with the model; it is not transmitted for each video. Each sequence-specific test caption is encoded with zstandard level 22 using this dictionary. We compare against Python zlib at level 9 and verify exact recovery after decompression for both methods. Table~\ref{tab:prompt-compression} reports the compressed sizes and compression ratios. To express textual side information on the same scale as the video rate, we amortize one sequence caption over a 96-frame $1920\times1080$ video. Dictionary-based zstandard reduces the average size by 39.6--42.5\% relative to zlib and limits the additional rate to $7.40\times10^{-6}$--$9.85\times10^{-6}$ bpp. Even relative to our lowest visual operating point of 0.0005 bpp, this overhead is only 1.5--2.0\%. The fixed general prompt is excluded from these values because it requires no per-video signaling.

\subsection{Discussion on Prompt}
\label{app:prompt-discussion}

To isolate the value of sequence-specific semantic side information, we compare the complete prompt against the fixed general quality prompt alone while keeping the codec and decoder unchanged. Table~\ref{tab:ablation:prompt} reports BD-rate relative to the corresponding general-only baseline on HEVC-B. Adding the sequence caption improves all four perceptual criteria for both decoding modes. UniGVC-BI reduces BD-rate by 1.88\% in LPIPS, 11.19\% in DISTS, 11.06\% in FID, and 17.11\% in KID. UniGVC-AR yields corresponding reductions of 2.64\%, 6.04\%, 14.29\%, and 16.11\%. The consistent improvements for both decoders indicate that the caption complements the shared codec representation rather than benefiting only one synthesis path. The larger gains in FID and KID than in LPIPS further suggest that sequence-level text primarily helps resolve semantic and distributional ambiguity at these extremely low rates.

\begin{table*}[t]
\caption{\textbf{Effect of sequence-specific captions on HEVC-B.} BD-rate (\%) is measured relative to the corresponding model using only the fixed general prompt. Lower is better; a negative value denotes a rate reduction at matched quality.}
\label{tab:ablation:prompt}
\centering
\setlength{\tabcolsep}{3pt}
\renewcommand{\arraystretch}{1.18}
\footnotesize
\begin{minipage}[t]{0.48\textwidth}
\centering
\begin{tabular}{lrrrr}
\toprule
\multicolumn{5}{c}{\textbf{(a) UniGVC-BI}} \\
\midrule
Prompt & LPIPS & DISTS & FID & KID \\
\midrule
General only & 0.00 & 0.00 & 0.00 & 0.00 \\
+ sequence & -1.88 & -11.19 & -11.06 & -17.11 \\
\bottomrule
\end{tabular}
\end{minipage}
\hfill
\begin{minipage}[t]{0.48\textwidth}
\centering
\begin{tabular}{lrrrr}
\toprule
\multicolumn{5}{c}{\textbf{(b) UniGVC-AR}} \\
\midrule
Prompt & LPIPS & DISTS & FID & KID \\
\midrule
General only & 0.00 & 0.00 & 0.00 & 0.00 \\
+ sequence & -2.64 & -6.04 & -14.29 & -16.11 \\
\bottomrule
\end{tabular}
\end{minipage}
\end{table*}

\begin{figure*}[t]
    \centering
    \includegraphics[width=\textwidth]{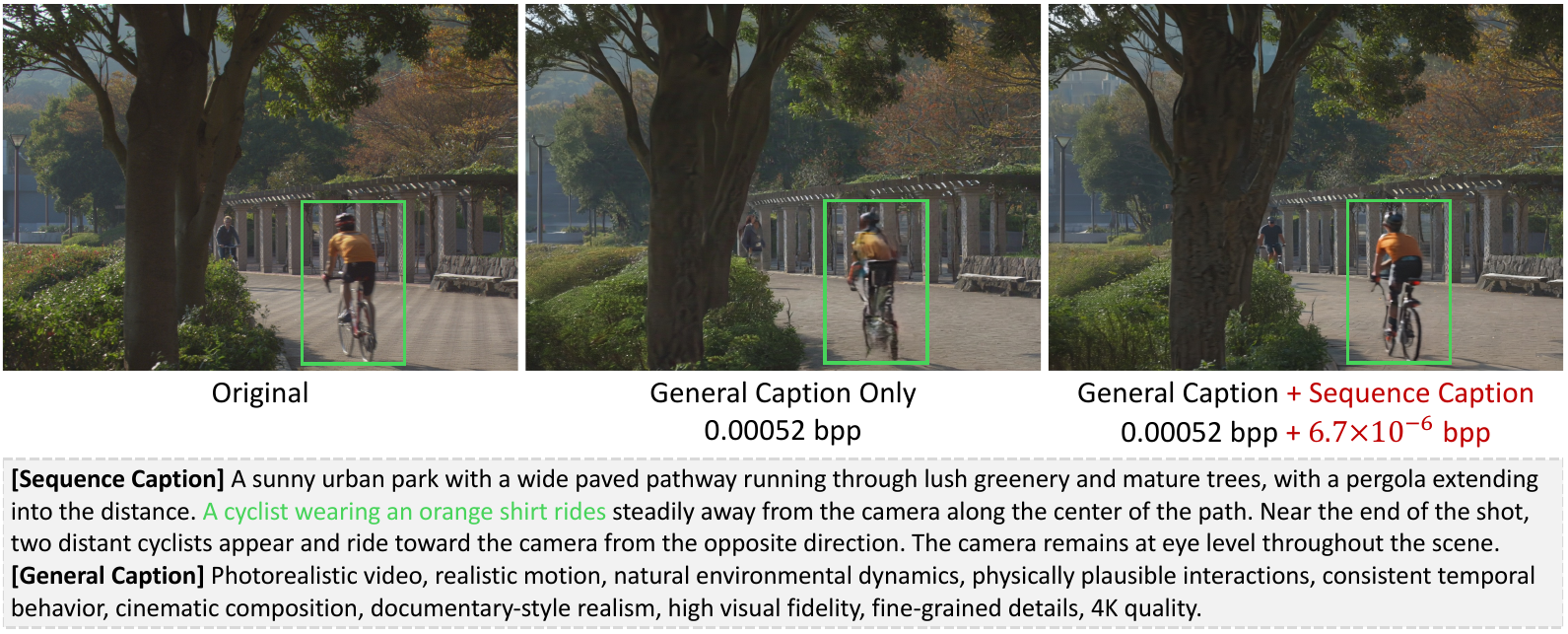}
    \caption{\textbf{Qualitative effect of the sequence-specific caption.} The columns show the source frame, UniGVC-BI's reconstruction using only the fixed general prompt, and reconstruction after adding the sequence caption. Both reconstructions use the same 0.00052-bpp bitstream; the sequence caption adds only $6.7\times10^{-6}$ bpp. The exact sequence and general prompts are shown below the examples.}
    \label{fig:prompt}
\end{figure*}

Fig.~\ref{fig:prompt} visualize the influence. With only the fixed general prompt, the decoder is informed of the desired visual style but receives no textual description of this particular scene. The sequence caption supplies the setting, subjects, motion, and camera behavior that are difficult to communicate through a 0.00052-bpp visual code alone. Adding it produces a reconstruction that more faithfully reflects the source content, while its compressed rate is only $6.7\times10^{-6}$ bpp for this example. Together with the quantitative gains in Table~\ref{tab:ablation:prompt}, this example supports using a concise sequence description as low-cost semantic side information rather than relying on a generic quality prompt alone.

\section{Additional Quantitative Comparison}
\label{app:additional-evaluation}

\begin{figure*}[t]
    \centering
    \includegraphics[width=\textwidth]{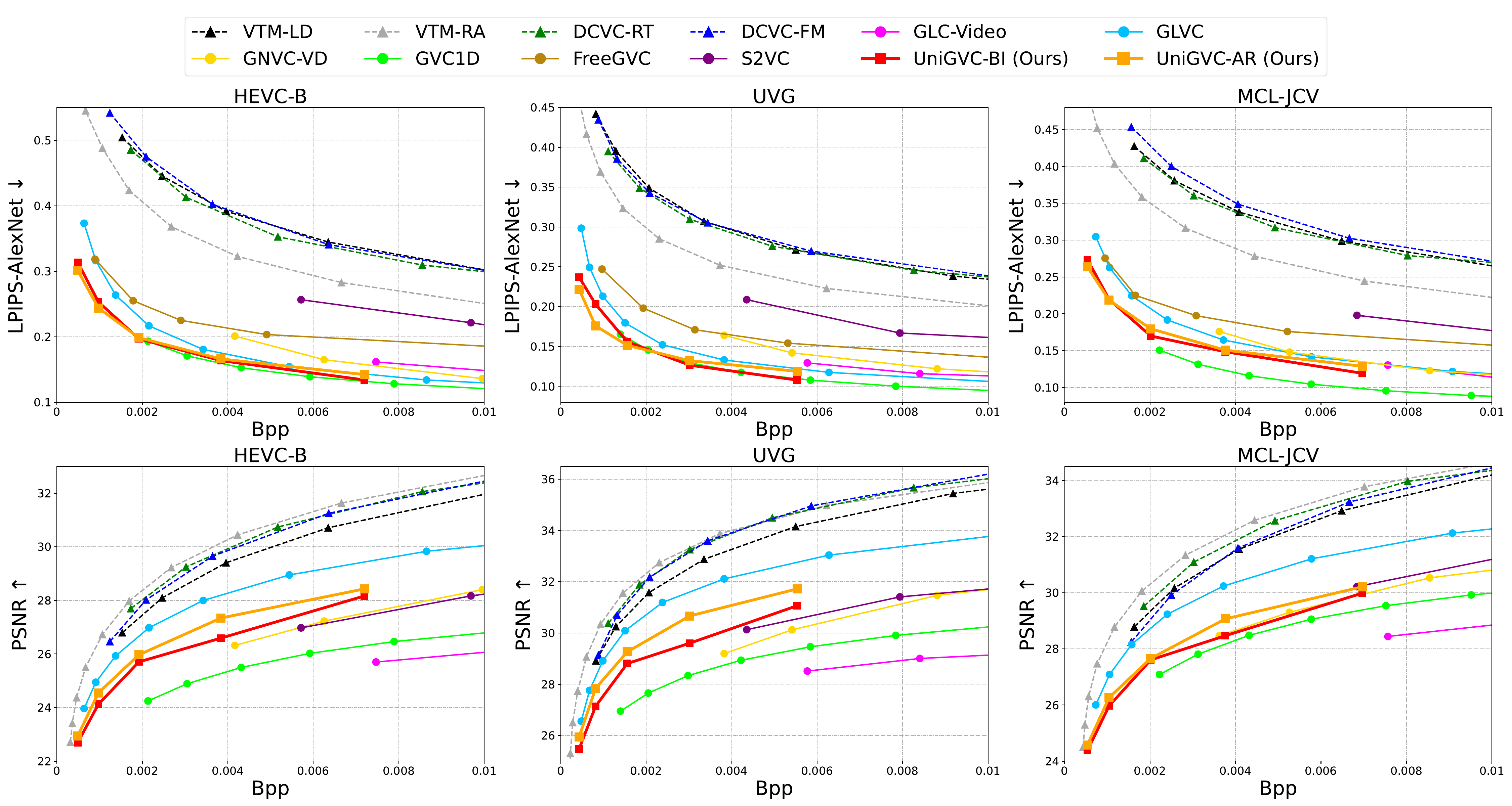}
    \caption{\textbf{Additional comparisons in terms of LPIPS-AlexNet and PSNR on HEVC-B, UVG, and MCL-JCV.} All methods follow the same 96-frame evaluation protocol as Fig.~\ref{fig:rdcurve}.}
    \label{fig:rdcurve_rest}
\end{figure*}

Fig.~\ref{fig:rdcurve_rest} complements the main comparison with LPIPS using an AlexNet backbone and with pixel-domain PSNR. The LPIPS-AlexNet results follow the perceptual trend observed with LPIPS-VGG in Fig.~\ref{fig:rdcurve}: UniGVC-BI and UniGVC-AR remain on or close to the favorable low-rate frontier across HEVC-B, UVG, and MCL-JCV. Pixel-level distortion metrics such as PSNR favor overly smooth or blurred reconstructions at ultra-low bitrates, which deviate substantially from human perception. We therefore report PSNR only as a supplementary distortion measure for reference, while evaluating ultra-low-rate compression primarily in terms of perceptual fidelity, temporal coherence, and human preference.

\section{Additional Qualitative Comparison}

\begin{figure*}[t]
    \centering
    \includegraphics[width=\textwidth]{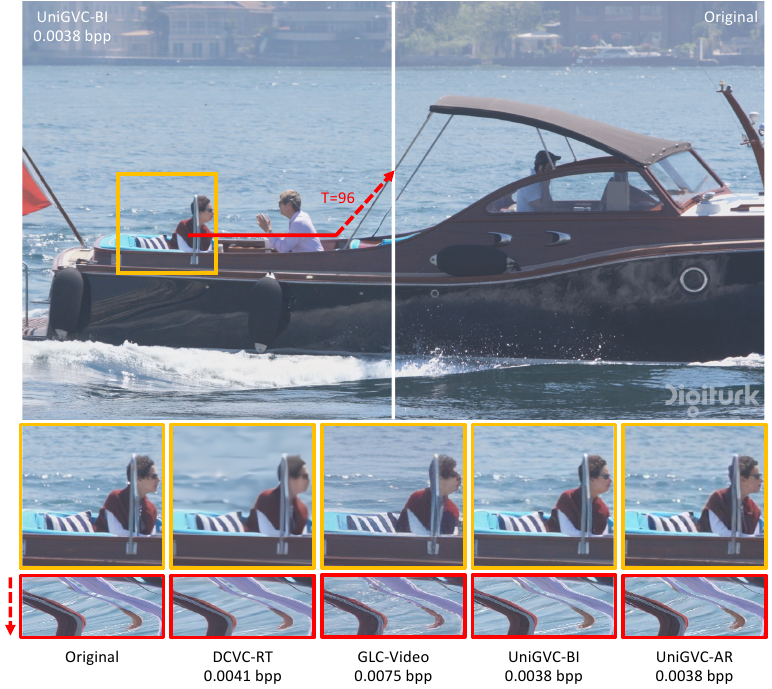}
    \caption{\textbf{Visual example from UVG.} \textit{Best viewed on screen for details.}}
    \label{fig:add_visual1}
\end{figure*}

\begin{figure*}[t]
    \centering
    \includegraphics[width=\textwidth]{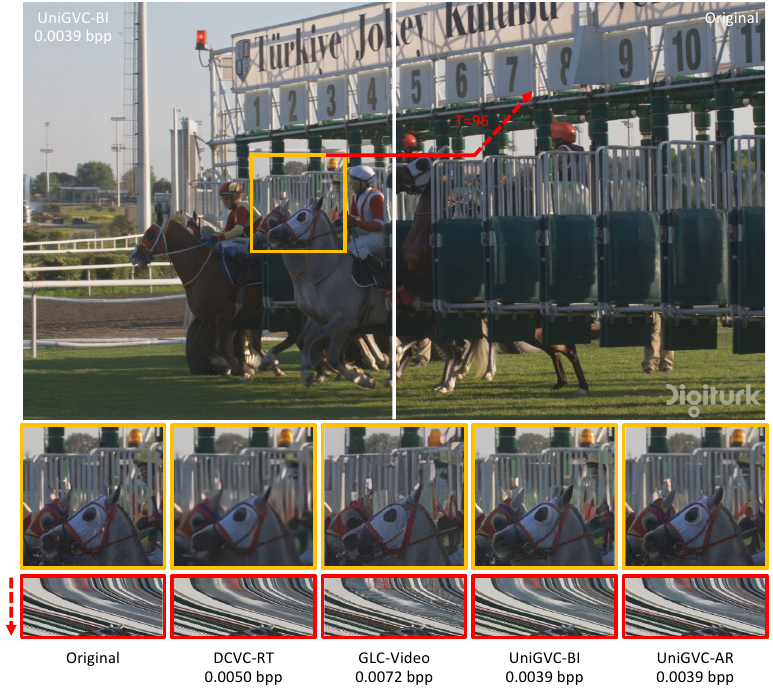}
    \caption{\textbf{Visual example from UVG.} \textit{Best viewed on screen for details.}}
    \label{fig:add_visual2}
\end{figure*}

\begin{figure*}[t]
    \centering
    \includegraphics[width=\textwidth]{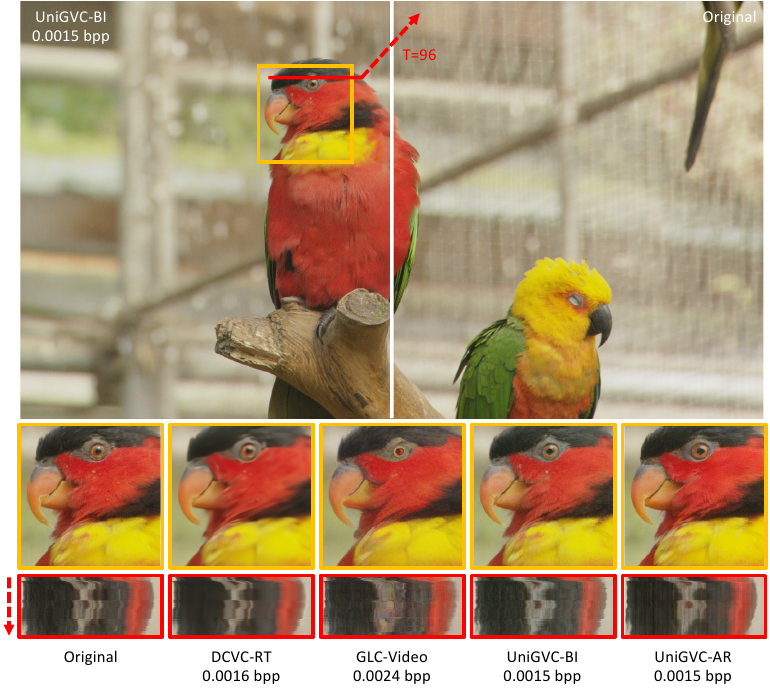}
    \caption{\textbf{Visual example from MCL-JCV.} \textit{Best viewed on screen for details.}}
    \label{fig:add_visual3}
\end{figure*}

\begin{figure*}[t]
    \centering
    \includegraphics[width=\textwidth]{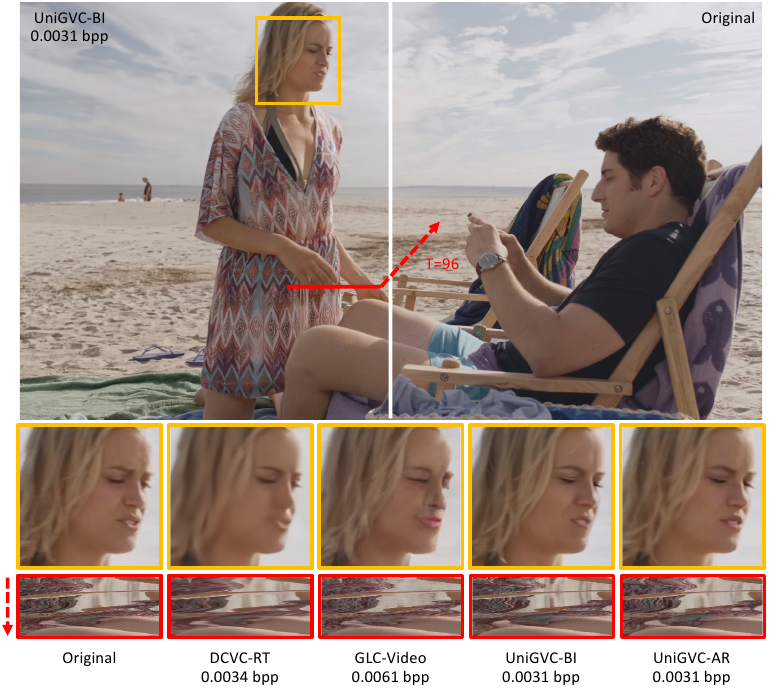}
    \caption{\textbf{Visual example from MCL-JCV.} \textit{Best viewed on screen for details.}}
    \label{fig:add_visual4}
\end{figure*}

Fig.~\ref{fig:add_visual1}--\ref{fig:add_visual4} provide additional qualitative comparisons on UVG and MCL-JCV. Each example presents reconstructed patches together with a 96-frame temporal profile, allowing spatial fidelity and temporal coherence to be assessed jointly. Both UniGVC-BI and UniGVC-AR deliver faithful, temporally consistent reconstructions while using fewer bits than GLC-Video and DCVC-RT.

\end{document}